\documentclass[journal]{IEEEtran}
\pdfoutput=1
\usepackage{booktabs}
\usepackage{indentfirst}
\usepackage{graphicx}
\usepackage{amsmath}
\usepackage{latexsym}
\usepackage{amssymb}
\usepackage{subfigure}
\usepackage{hyperref}
\usepackage{cite}
\usepackage{color}
\usepackage{stfloats}
\usepackage{multirow}
\usepackage[lined,ruled,boxed]{algorithm2e}

\begin{document}
%
\title{Discriminative Nonlinear Analysis Operator Learning: When Cosparse Model Meets Image Classification}
%
%
\author{Zaidao Wen,~Biao Hou,~\emph{Member,~IEEE}, and~Licheng Jiao,~\emph{Senior Member,~IEEE}}

\maketitle

\begin{abstract}
Linear synthesis model based dictionary learning framework has achieved remarkable performances in image classification in the last decade. Behaved as a generative feature model, it however suffers from some intrinsic deficiencies. In this paper, we propose a novel parametric nonlinear analysis cosparse model (NACM) with which a unique feature vector will be much more efficiently extracted. Additionally, we derive a deep insight to demonstrate that NACM is capable of simultaneously learning the task adapted feature transformation and regularization to encode our preferences, domain prior knowledge and task oriented supervised information into the features. The proposed NACM is devoted to the classification task as a discriminative feature model and yield a novel discriminative nonlinear analysis operator learning framework (DNAOL). The theoretical analysis and experimental performances clearly demonstrate that DNAOL will not only achieve the better or at least competitive classification accuracies than the state-of-the-art algorithms but it can also dramatically reduce the time complexities in both training and testing phases.
\end{abstract}

\begin{keywords}
Nonlinear analysis cosparse model, analysis operator learning, regularization learning, generative model, discriminative model, dictionary learning, linear synthesis model, image classification.
\end{keywords}

%
\IEEEpeerreviewmaketitle

\section{Introduction}\label{Sec:Introduction}

\PARstart{I}{mage} classification is one of the important tasks in the field of image processing, computer vision, and pattern recognition. To implement this task, let $\{\mathbf{x}_i\}_{i=1}^N$ be a collection of $N$ training samples drawn from $C$ classes with labels $\{\mathbf{y}_i\}_{i=1}^N$. Supervised classification method will predict the label vector $\mathbf{\widehat{y}}$ for a query test sample $\mathbf{\widehat{x}}$ based on the training pairs $\{\mathbf{x}_i,\mathbf{y}_i\}_{i=1}^N$, which can be formulated as the following maximum a posterior (MAP) problem from the viewpoint of the probabilistic model \cite{Bishop2006}:
\begin{equation}\label{Equ:Classification}
  \max_{\mathbf{\widehat{y}}} \mathbb{P}(\mathbf{\widehat{y}}|\widehat{\mathbf{x}},\{\mathbf{x}_i,~\mathbf{y}_i\}_{i=1}^N)
\end{equation}
In general, this problem will be addressed with the help of two sequential procedures, i.e., feature extraction and classifier construction. Instead of the conventional feature engine fashion where a large amount of features are collected from some user-specific non-parametric transformations according to the domain knowledge \cite{lazebnik2006beyond,Mairal2012,bengio2013representation,liu2002gabor,lowe2004distinctive,dalal2005histograms}, the learning based parametric models are much appealing in recent year. These methods aim at computing a parametric feature extraction operator $\mathcal{P}$ associated with some more discriminative and representative features $\{\mathbf{f}_i\}_{i=1}^N$ as well as a classifier $\mathcal{W}$ in a joint framework, yielding the following approximation classification schemes \cite{Bernardo2007}. Here and after, we use the matrix notation to indicate a set of vectors for clarity, \emph{e.g.}, $\mathbf{X}=\{\mathbf{x}_i\}_{i=1}^N$.
\begin{equation}\label{Equ:Classification_parametric}
\begin{split}
\setlength{\abovedisplayskip}{1pt}
\arg\max_{\mathbf{\widehat{y}},\mathbf{\widehat{f}}} \mathbb{P}(\mathbf{\widehat{y}},\mathbf{\widehat{f}}|\widehat{\mathbf{x}},\mathbf{X},\mathbf{Y})\approx \arg\max_{\mathbf{\widehat{y}},\mathbf{\widehat{f}}} \mathbb{P}(\mathbf{\widehat{y}},\mathbf{\widehat{f}}|\widehat{\mathbf{x}},\theta_{\mathcal{P}}^*,\theta_{\mathcal{W}}^*)\\
=\arg\max_{\mathbf{\widehat{y}},\mathbf{\widehat{f}}} \mathbb{P}(\mathbf{\widehat{y}},\widehat{\mathbf{x}},\widehat{\mathbf{f}}|\theta_{\mathcal{P}}^*,\theta_{\mathcal{W}}^*)\\
 \mathrm{s.t.}~\{\theta_{\mathcal{P}}^*,\theta_{\mathcal{W}}^*,\mathbf{F}\}=\arg\max_{\theta_{\mathcal{P}},\theta_{\mathcal{W}},\mathbf{F}} \mathbb{P}(\theta_{\mathcal{P}},\theta_{\mathcal{W}},\mathbf{Y,F}|\mathbf{X})\\
 =\arg\max_{\theta_{\mathcal{P}},\theta_{\mathcal{W}},\mathbf{F}} \mathbb{P}(\mathbf{X,Y,F},\theta_{\mathcal{P}},\theta_{\mathcal{W}})
 \end{split}
\end{equation}
where $\mathbf{\widehat{f}}$ is the feature vector for the query sample, $\theta_{\mathcal{P}},\theta_{\mathcal{W}}$ stand for the parameters for $\mathcal{P}$ and $\mathcal{W}$, respectively and $\theta_{\mathcal{P}}^*,\theta_{\mathcal{W}}^*$ will be the MAP estimations of the problem in the constraint corresponding to the training phase. It can be observed from \eqref{Equ:Classification_parametric} that these optimal MAP solutions will be obtained by maximizing either the posterior distribution or joint distribution following the Bayes' law. As a consequence, the core issue will become to characterize the joint or posterior distribution with a suitable parametric model.
\subsection{Regularized Linear Synthesis Model}\label{subsec:DLSSM}
\par In the last decade, the regularized linear synthesis model attracts so many attentions that it has achieved remarkable progresses in many image processing applications such as image de-noising and super-resolution \cite{rubinstein2010dictionaries,Elad2010,aharon2006svd,yang2010image}. As a parametric model, it characterizes an input sample $\mathbf{x}$ via the linear combinations $\mathbf{x}\approx\mathbf{Df}$, where $\mathbf{D}\in\mathbb{R}^{n\times p}$ is referred to as the dictionary formed of $p$ column vectors termed atoms and the coefficient vector $\mathbf{f}$ is a regularized hidden feature vector of $\mathbf{x}$. Specially, this model in statistical machine learning is also known as linear regression with a distinct motivations and explanations \cite{seber2012linear,Mairal2014}. In above image processing applications, one always focuses on optimizing the column vectors, namely atoms in $\mathbf{D}$ to achieve a minimized representation error, which is generally called synthesis dictionary learning (DL) \cite{aharon2006svd,rubinstein2010dictionaries}. On the contrary, we will pay more attention to the properties of $\mathbf{f}$ in some machine learning tasks \cite{Wright2009}. When this model meets the task of image classification, roughly speaking, it appears as a generative model to characterize the joint distribution in \eqref{Equ:Classification_parametric} with the following two types of classification implementations:
\begin{equation}\label{Equ_Twotype_model}
\begin{split}
  \max_{\theta_{\mathcal{P}},\theta_{\mathcal{W}},\mathbf{F}} \mathbb{P}(\mathbf{X,Y,F},\theta_{\mathcal{P}},\theta_{\mathcal{W}})
  \\
  \overset{\text{(1)}}{=}\max_{\theta_{\mathcal{P}},\theta_{\mathcal{W}},\mathbf{F}} \mathbb{P}(\mathbf{Y}|\mathbf{X,F},\theta_{\mathcal{W}})\mathbb{P}(\theta_{\mathcal{P}},\theta_{\mathcal{W}})
  \mathbb{P}(\mathbf{X}|\mathbf{F},\theta_{\mathcal{P}})\mathbb{P}(\mathbf{F})\\
  \overset{\text{(2)}}{=}\max_{\theta_{\mathcal{P}},\theta_{\mathcal{W}},\mathbf{F}} \mathbb{P}(\mathbf{Y}|\mathbf{F},\theta_{\mathcal{W}})\mathbb{P}(\theta_{\mathcal{W}})p(\theta_{\mathcal{P}})  \mathbb{P}(\mathbf{X}|\mathbf{F},\theta_{\mathcal{P}})\mathbb{P}(\mathbf{F})
  \end{split}
\end{equation}
\par Considering the first type, the parametric classifier model $\mathbb{P}(\mathbf{Y}|\mathbf{X,F},\theta_{\mathcal{W}})$ will learn a set of $\theta_{\mathcal{W}}=\{\mathbf{W}^c\}_{c=1}^C$ in the spirit of regression such that each $\mathbf{W}^c$ will mostly fit the samples from class $c$ \cite{ramirez2010classification,kong2012dictionary,yang2014sparse,Wen2016}. Accordingly, classification can be realized by measuring the regression residuals. Alternatively, the second type of classifier will focus on $\mathbb{P}(\mathbf{Y}|\mathbf{F},\theta_{\mathcal{W}})$ to measure the classification loss in a straightforward way, where some off-the-shelf loss functions in the typical classifiers will be exploited to model this term, \emph{e.g.}, softmax, square $\ell_2$ norm in ridge regression \cite{Mairal2009,zhang2010discriminative,jiang2013label,Zhang2016,Jiang2016,Zhang2017}. The rest $\mathbb{P}(\theta_{\mathcal{W}})$ will correspond to a regularization on the classifier parameters $\theta_{\mathcal{W}}$ to prevent from over-fitting. More detailed information for these two types of classifiers will be discussed in Sec. \ref{subsec:framework} and this paper will focus on the latter two shared terms corresponding to the feature model.
\subsection{Motivation}\label{Subsec:Motivation}

\par In the previous researches, the parametric feature model $\mathbb{P}(\mathbf{X}|\mathbf{F},\theta_{\mathcal{P}})\mathbb{P}(\mathbf{F})$ will be characterized by the above regularized linear synthesis model with $\theta_{\mathcal{P}}=\mathbf{D}$ and a regularization term on $\mathbf{F}$, \emph{e.g.,} sparsity inducing $\ell_1$ norm \cite{Wright2009,Yang2014Latent,ramirez2010classification,kong2012dictionary}, square $\ell_2$ norm \cite{zhang2011sparse}, group sparsity inducing $\ell_{1,2}$ norm \cite{Sun2014}, low rank inducing nuclear norm \cite{zhang2013learning}, their combinations \cite{Zhang2013}, hierarchical prior \cite{Wen2016}, elastic net and Fisher term \cite{Mairal2012,yang2014sparse}. Although these different regularizations can all compute the MAP solutions of a discriminative dictionary and features to promote the classification performance, this feature model will suffer from the following intrinsic problems.
\par Let us firstly consider the term $\mathbb{P}(\mathbf{X}|\mathbf{F},\mathbf{D})$. When $\mathrm{rank}(\mathbf{D})<p$ for either an over-complete or an under-complete compact dictionary with many coherent atoms, $\mathbf{D}$ will exist a non-trivial null space. In this case, there will be generally a large amount of candidate feature vectors fitting this term while only a few of them will be of benefit to the classification task. As a consequence, once an incorrect atom is selected during feature extraction in training or testing phases, it will subsequently result in a domino effect to yield an incorrect classification result or training performance \cite{Elad2007}. Although many strategies have been proposed to relieve this problem \cite{Wen2016,kong2012dictionary,ramirez2010classification}, it will be always an intrinsic deficiency caused by the synthesis formulation, \emph{i.e.}, $\mathbf{x}\approx \mathbf{Df}$.
\par Secondly, in order to describe the term $\mathbb{P}(\mathbf{F})$, various user-specific regularizations have been exploited to encode the domain prior knowledge and our preferences in features to make a selection, resulting in numerous supervised or semi-supervised DL frameworks \cite{Cai2014,Mairal2009,zhang2011sparse,ramirez2010classification,elhamifar2011robust,Zhang2011,kong2012dictionary,Mairal2012,jiang2013label,wang2013max,Zhang2013,zhang2013learning,Sun2014,Yang2014Latent,yang2014sparse,Jiang2016,Wen2016,Zhang2016,Zhang2017}. Since no explicit discriminative strategy is encoded in the most regularizers, whether these regularizations or selections will be benefit to the task at hand and fit for the input samples is still an open question \cite{zhang2011sparse} whereas tuning their adjunctive regularization parameters will bring us another tedious problem to achieve a satisfactory generalizable performance. What's worth, most of these regularizers are non-smoothness so that the feature extractor $\mathcal{P}$ will be an inexplicit complex nonlinear function without the closed form. It follows that it is rather time consuming for feature extraction in both training and testing phases by means of some iterative optimization algorithms \cite{tropp2007signal,daubechies2004iterative,Wen2017}, which will heavily limit its applications to some large scale problems. Some efforts have been made for acceleration, including fast recovery algorithms \cite{Beck2009,lee2006efficient}, training a parametric function with the explicit form to approximate $\mathcal{P}$  \cite{gregor2010learning,szlam2012fast,Gu2014,Zhang2016}, introducing the online strategy \cite{mairal2010online}, \emph{etc}, but this intrinsic inefficiency caused by the regularized linear synthesis model cannot be eliminated. Can we directly learn a task specific parametric prior term to develop a task-driven prior learning framework following the way of dictionary learning \cite{Wen2016,Mairal2012}?
\par Finally, since feature vectors are coupled in the classifier and feature model in \eqref{Equ_Twotype_model}, both types of implementations will suffer from the inconsonant feature extractors in training and testing phases \cite{Mairal2014}. More concretely, since the label is not available in the testing phase, it can be concluded that the inexplicit feature extractor will be $\mathbf{f}=\mathcal{P}_{\mathrm{train}}(\mathbf{x|\theta_{\mathcal{P}}^*,y,\theta_{\mathcal{W}}^*})$ and $\mathbf{\widehat{f}}=\mathcal{P}_{\mathrm{test}}(\mathbf{\widehat{x}|\theta_{\mathcal{P}}^*})$ in the training and testing phases, respectively. To our best knowledge, this problem is only concerned and addressed in \cite{Mairal2012} at the cost of solving a relatively complicated bi-level optimization.
\subsection{Main Objective and Contributions}
\par According to the above motivations, the mentioned deficiencies are mostly caused by modeling the joint distribution with the regularized linear synthesis model. As a consequence, the objective of this paper is to solve the problems raised above by developing a parametric discriminative model to characterize the alternative counterpart in \eqref{Equ:Classification_parametric}, namely the posterior distribution. Our main contributions can be summarized as follows:
\begin{itemize}
  \item We propose a novel parametric nonlinear analysis cosparse model (NACM) with which a unique MAP solution of feature vector will be much more efficiently extracted than the conventional regularized linear synthesis model.
  \item We derive a deep insight to demonstrate that NACM is capable of simultaneously learning the task adapted feature transformation and regularization to encode our preferences, domain prior knowledge and task oriented supervised information into the features.
  \item We develop a discriminative nonlinear analysis operator learning framework (DNAOL) to devote the cosparse model to the classification task, in which the feature extractor is consistently determined by NACM in both training and testing phases.
\end{itemize}
\par Our proposed framework is validated on four image benchmarks for classification. By evaluating the classification accuracy and time costs in the training and testing phases, our framework can achieve higher accuracy than their counterpart dictionary learning algorithms with the same discriminative strategies. Compared with other state-of-the-art DL algorithms containing more discriminative strategies, our framework can still achieve better or competitive classification accuracies while the required time costs will be dramatically reduced, which evidently demonstrates the effectiveness and superiorities of our proposed framework. The rest paper is organized as follows. In Sec. \ref{Sec:Relatedwork}, we present a preliminary on linear analysis cosparse model and the related researches on analysis operator learning. We propose NACM and discuss its superiority in Sec. \ref{Sec:ModelPresentation}. In Sec. \ref{Sec:DPCOL}, we develop a DPNOL framework with two classification schemes and we derive a detailed optimization algorithm for parameters learning. We evaluate the proposed framework on several popular image databases in Sec. \ref{Sec:Experiment} and concluding remarks are given in Sec. \ref{Sec:Concluding}. For clarity, the important notations are summarized in the Table \ref{Tab:Notation}.
\begin{table}  \centering  \caption{Notations}\label{Tab:Notation}
  \begin{tabular}{|c|c|}
\hline
 notation& interpretation  \tabularnewline
\hline
\hline
 $\mathbf{x}_i \in\mathcal{X}^n$&$i$-th sample from the $n$-dimensional input domain  \tabularnewline
\hline
 $\mathcal{F}^p$ & $p$-dimensional feature domain \tabularnewline
 \hline
  $\mathcal{P}:\mathcal{X}\mapsto\mathcal{F}$ & feature extraction mapping (extractor) \tabularnewline
\hline
 $\theta_{\mathcal{P}}$, $\theta_{\mathcal{W}}$  & parameters in feature model and classifier \tabularnewline
 \hline
 $\mathbf{y}_i\in\mathcal{L}$ & label vector of $\mathbf{x}_i$ in the label domain \tabularnewline
 \hline
 $\mathbf{D}$, $\mathbf{D}_{:,i}$ & synthesis dictionary and its $i$-th column vector \tabularnewline
  \hline
 $\mathbf{A}$, $\mathbf{A}_{i,:}$  & Analysis operator and its $i$-th row vector \tabularnewline
    \hline
 $\mathbf{f}_i$, $\mathbf{f}_i(j)$ & $i$-th feature vector and its $j$-th entry  \tabularnewline
    \hline
 $\mathbf{X}^c$, $\theta_{\mathcal{W}}^c$, $\theta_{\mathcal{P}}^c$  & variables with respect to  class $c$  \tabularnewline
     \hline
 $<\mathbf{A,B}>$ & the inner product between matrix $\mathbf{A}$ and $\mathbf{B}$  \tabularnewline
      \hline
 $\overline{\mathbf{X}}^c$ & sample matrix excluding c-th class  \tabularnewline
       \hline
 $\mathcal{N}$, $\mathcal{U}$ & normal and uniform distribution  \tabularnewline
     \hline
      $\mathbb{P}(\cdot)$ & probability density function  \tabularnewline
      \hline
$\mathbf{1}$,$\mathbf{0}$ & full 1 and 0 vector, respectively  \tabularnewline
      \hline
      $\mathbf{I}$ & identity matrix  \tabularnewline
      \hline
\end{tabular}
\end{table}

\section{Preliminary}\label{Sec:Relatedwork}
This section will introduce the conventional linear analysis model and the related researches on analysis operator learning (AOL).
\subsection{Linear Analysis Model}
\par As aforementioned, the regularized linear synthesis model describes a signal $\mathbf{x}\in\mathbb{R}^n$ with the linear combinations of some atoms in $\mathbf{D}$, which implies $\mathbf{x}$ is generated from $\mathbf{D}$ and a regularized $\mathbf{f}$. Alternatively, another counterpart termed linear analysis model takes a different viewpoint by characterizing $\mathbf{x}$ with a collection of linear filters in the rows of $\mathbf{A}\in\mathbb{R}^{p\times n}$ as $\mathbf{f=Ax}$, where $\mathbf{f}$ is called the analysis feature vector with respect to the analysis operator $\mathbf{A}$ \cite{Elad2007}. More concretely, linear analysis model will resolve or analyze $\mathbf{x}$ with a set of filters in $\mathbf{A}$. It follows that when the analysis operator $\mathbf{A}$ and the synthesis dictionary $\mathbf{D}$ are both square nonsingular matrices and $\mathbf{A}^{-1}=\mathbf{D}$, these two models will be identical during characterizing $\mathbf{x}$ as $\mathbf{x=Df=DAx}$. Attentions have been recently attracted by a special instance referred to as analysis cosparse model (ACM) in which the analysis feature vector $\mathbf{f=Ax}$ will contain many zero components \cite{Nam2013}. In this situation, $\mathbf{x}$ will be confined to a polyhedral of the null space of a sub-row matrix corresponding to those filters with zero responses, namely $\mathbf{f}(j)=\mathbf{A}_{j,:}\mathbf{x}=0$. Accordingly, more attentions will be paid on these row vectors in this specific model and their amounts will be defined as the cosparsity of $\mathbf{x}$ with respect to $\mathbf{A}$ \cite{Nam2013}, contrasting the concept of the sparsity in the synthesis model \cite{Elad2010}.
\subsection{Analysis Operator Learning}
\par Analogous to the early stage of DL with the regularized linear synthesis model, the pioneering researches of ACM will mostly work on some basic image inverse problems \cite{Elad2007}, such as de-noising, de-bluring and super-resolution, where the analysis operator $\mathbf{A}$ will be adaptively learned to meet the requirement of these recovery tasks \cite{Hawe2013}. As a straightforward extension of K-Singular value decomposition (K-SVD) \cite{aharon2006svd}, Rubinstein \emph{et al.} presented an analysis K-SVD algorithm (AK-SVD) for AOL \cite{rubinstein2013analysis}. In \cite{yaghoobi2013constrained} and \cite{Hawe2013}, the authors restricted the operator on an oblique manifold or a uniform normalized tight frame to get rid of some trivial solutions. Additionally, several frameworks of learning a sparsifying transform have been developed \cite{Ravishankar2013,strcturedsparsifying2014}, which will benefit from the efficiency promotion during recovery. More recently, Bian \emph{et al.} investigated the relationship between the sparse null space problem and AOL, and presented an efficient learning algorithm based on the sparse null space basis pursuit scheme \cite{Bian2016}.
\par When this parametric model meets the classification task, to our best knowledge, there are far fewer researches focusing on learning a discriminative analysis operator than its counterpart. In \cite{shekhar2014analysis}, the authors presented a classification framework with ACM, where the SVM classifier is trained on the resulted cosparse feature vectors \cite{chang2011libsvm}. Nonetheless, no supervised information is encoded in the operator as well as features during parameters learning so that their performances are less optimal compared with the other state-of-the-art DL algorithms. Fawzi \emph{et al.} developed a discriminative AOL framework for fast classification \cite{fawzi2014dictionary}, in which the hinge loss function in SVM is incorporated in the AOL framework to enhance the discrimination of the operator. Additionally, Gu \emph{et al}. presented a dictionary pair learning (DPL) framework to simultaneously learn a synthesis dictionary and an analysis operator for discriminative classification \cite{Gu2014}.
\section{Nonlinear Analysis Based Cosparse Model}\label{Sec:ModelPresentation}
In this section, we will develop the novel parametric nonlinear analysis cosparse model (NACM) and present a discussion to demonstrate its superiorities and novelties.
\subsection{From Linear to Nonlinear Model}
\par According to the above preliminary, ACM will produce the unique cosparse feature vector in a deterministic way as $\mathbf{f=Ax}$, where only a matrix-vector multiplication operation is required. It follows that this model will be much more efficient for feature extraction than its counterpart, namely regularized linear synthesis model. Nevertheless, deficiency comes from the fact that the linear analysis operator $\mathbf{A}$ will not increase the separability of input samples while the sparse pattern, namely indices corresponding to the zero feature response will be sensitive to the perturbation. Inspired by the kernel method \cite{chang2011libsvm}\cite{Hofmann2008}, a novel nonlinear analysis cosparse model (NACM) formulated as $\mathbf{f}=\mathcal{P}(\mathbf{x})=\mathcal{S}_{\mathbf{\Lambda}}(\mathbf{A}\mathbf{x})$ is extended from the conventional linear ACM, where a nonlinear cosparse operator $\mathcal{P}$ is explicitly determined by the model parameters $\theta_{\mathcal{P}}=\{\mathbf{A},\mathbf{\Lambda}\}$ and $\mathcal{S}_{\mathbf{\Lambda}}$ is a parametric nonlinear function generating a cosparse output vector\footnote{Since our model is a straightforward extension from ACM, all concepts therein will be also inherited in our model, \emph{e.g.,} cosparse feature vector, cosparsity, \emph{etc}.}. With this parametric nonlinear operator, it is possible to incorporate some supervised information to enhance the discrimination of the cosparse feature vectors. Note that if $\mathcal{S}_{\mathbf{\Lambda}}$ is further assumed to be separable with respect to each component, each feature will be also efficiently computed as $\mathbf{f}(j)=\mathcal{S}_{\mathbf{\Lambda}}(\mathbf{A}_{j,:} \mathbf{x}),~j=1,\dots,p$ allowing a fast distributed parallel computing. In the following part, we will derive a deep insight to give an interpretation of its formulation and demonstrate its superiorities.

\subsection{Model Interpretations and Superiorities}
Considering the formulation of NACM, if we refer to $\mathbf{A}$ and $\mathcal{S}_{\mathbf{\Lambda}}$ as a linear feature transformation and a nonlinear selection operator, respectively,  NACM will enable us to learn an explicitly discriminative feature transformation and selection by optimizing $\mathbf{A}$ and $\mathbf{\Lambda}$ adaptively according to the task at hand. In order to take a deeper insight, let us consider the following regularized optimization problem:
\begin{equation}\label{Equ:PCR_model}
  \min_{\mathbf{f}}\lambda \Omega(\mathbf{f})+ \frac{1}{2}\|\mathbf{f}-\mathbf{A}\mathbf{x}\|_{2}^2,
\end{equation}
where $\Omega(\mathbf{f})$ is a proper, convex, lower-semicontinuous regularization function on features with regularization hyper-parameter $\lambda$. This optimization can be regarded as a de-noised ACM framework in the feature domain, which aims to obtain a regularized MAP solution of $\mathbf{f}$ in the neighbourhood of $\mathbf{Ax}$ in the sense of Euclidean distance. In particular, when $\Omega$ is the $\ell_1$ norm and $\mathbf{A}$ is a square matrix, it will become the recently developed sparsifying transforms model \cite{Ravishankar2013}. In particular, the minimizer to \eqref{Equ:PCR_model} admits a closed form given by:
\begin{equation}\label{Equ:Solu_PCR}
  {\mathbf{f}^*}=\mathrm{prox}_{\Omega,\lambda}(\mathbf{Ax}),
\end{equation}
where $\mathrm{prox}_{\Omega,\lambda}$ is called the proximity operator of $\Omega$ with parameter $\lambda$ \cite{combettes2005signal}. Comparing \eqref{Equ:Solu_PCR} with the formulation of NACM, namely $\mathbf{f}=\mathcal{S}_{\mathbf{\Lambda}}(\mathbf{A}\mathbf{x})$, if $\mathcal{S}_{\mathbf{\Lambda}}$ is coincidently a proximity operator or its functional approximation of the regularizer $\Omega$, NACM will precisely generate a feature vector containing the regularized information of $\Omega$ such as prior knowledge and the preference. In other words, the performance of a feature selector $\mathcal{S}_{\mathbf{\Lambda}}$ will be equivalent to imposing an underlying regularization on cosparse features. It follows that the regularized features will be efficiently computed with an explicit formulation without exploiting extra iterative algorithms and it is thus much more convenient to impose various regularizations than the regularized linear synthesis model. In this spirit, directly optimizing the parameters in NACM in a task-driven manner will be interpreted as learning a task adapted feature transformation and regularization, which will be of more novelty and superiority than regularized synthesis model. In our previous research, we imposed a hierarchical prior on the features by leveraging a synthesis model for discriminative transformation and regularization learning \cite{Wen2016}, but its complexity will be much aggravated than that of NACM. Furthermore, NACM will essentially generate a series of MAP solution of $\mathbf{f}$ by varying model parameters and as a consequence, $\mathbf{f}^*=\mathcal{S}_{\mathbf{\Lambda}}(\mathbf{Ax})$ actually provides a lower bound of the posterior distribution $\max_{\theta_{\mathcal{P}}}\mathbb{P}(\mathbf{f}^*,\theta_{\mathcal{P}}|\mathbf{x})$ as:
\begin{equation}\label{Equ:Lowerbound}
  \max_{\mathbf{f},\theta_{\mathcal{P}}}\mathbb{P}(\mathbf{f,\theta_{\mathcal{P}}|x})\leq \max_{\mathbf{\theta_{\mathcal{P}}}}\mathbb{P}(\mathbf{f}^*,\mathbf{\theta_{\mathcal{P}}|x})\leq \mathbb{P}(\mathbf{f}^{**},\theta_{\mathcal{P}}^*|\mathbf{x})
\end{equation}
where $\mathbf{f}^{**}$ and $\theta_{\mathcal{P}}^*$ stand for the MAP solutions of features and parameters, respectively.
\par Compared with the conventional ACM, the proposed NACM interpreted as a regularized variant will focus more on the properties of the feature domain while ACM normally provides a prior information of the input domain \cite{Elad2007}. Additionally, the parametric NACM has a much more powerful capability of functional approximation than its linear counterpart, which can generalize its potential applications, \emph{e.g.}, nonlinear dimensionality reduction in the case of $p<n$. Compared with the kernel method, both of them exploit the nonlinear mapping to enhance the discrimination for a set of input samples without greatly increase the model complexity, but NACM will moreover impose regularizations on the explicit features. Therefore, NACM can be tailored for much broader problems requiring the explicit form of features while kernel tricks can be only adapted to problems with some specific forms.


\section{Discriminative Nonlinear Analysis Operator Learning}\label{Sec:DPCOL}

\par In this section, the proposed NACM is formally adapted to the classification task to develop a discriminative framework termed DNAOL to model the posterior distribution in \eqref{Equ_Twotype_model}, in which the parametric nonlinear analysis operator will be learned to meet the requirement of the classification task. In order to demonstrate the advantages of NACM during constructing the feature model over the conventional regularized linear synthesis model, the terms with respect to the classifier in \eqref{Equ_Twotype_model} should be inherited for fairness without involving more discriminative strategies. For this reason, we also present two types of classification schemes in a corresponding manner. Then we derive a detailed optimization scheme for framework learning. We conclude this section by theoretically comparing our framework with other state-of-the-art DL frameworks and AOL methods to highlight the superiorities.
\subsection{Framework Proposition}\label{subsec:framework}
\par In the spirit of two schemes in \eqref{Equ_Twotype_model} described by the generative model, we will also present two types of implementations to characterize the posterior distribution with the following problem:
\begin{equation}\label{Equ:Classification_parametricposterior}
\begin{split}
\max_{\mathbf{\widehat{y}},\mathbf{\widehat{f}}} \mathbb{P}(\mathbf{\widehat{y}},\mathbf{\widehat{f}}|\widehat{\mathbf{x}},\theta_{\mathcal{P}}^*,\theta_{\mathcal{W}}^*)\\
 \mathrm{s.t.}~\{\theta_{\mathcal{P}}^*,\theta_{\mathcal{W}}^*,\mathbf{F}\}=\arg\max_{\theta_{\mathcal{P}},\theta_{\mathcal{W}},\mathbf{F}} \mathbb{P}(\theta_{\mathcal{P}},\theta_{\mathcal{W}},\mathbf{Y,F}|\mathbf{X})\\
  \overset{\text{(1)}}{=}\mathbb{P}(\mathbf{Y}|\mathbf{X,F},\theta_{\mathcal{W}})\mathbb{P}(\theta_{\mathcal{W}})\mathbb{P}(\mathbf{F}|\mathbf{X},\theta_{\mathcal{P}})\mathbb{P}(\theta_{\mathcal{P}})\\
 \overset{\text{(2)}}{=}\mathbb{P}(\mathbf{Y}|\mathbf{X},\theta_{\mathcal{W}})\mathbb{P}(\theta_{\mathcal{W}})\mathbb{P}(\mathbf{F}|\mathbf{X},\theta_{\mathcal{P}})\mathbb{P}(\theta_{\mathcal{P}}).
 \end{split}
\end{equation}
However, we may note that the feature vectors are still coupled in the classifier and feature model in \eqref{Equ:Classification_parametricposterior} for both schemes, which will also yield the inconsonant problem mentioned in our motivation. To solve this problem, we consider to decouple two models by developing the following bi-level optimization in the training phase:
\begin{equation}\label{Equ:Classification_parametricposteriorrelaxition1}
\begin{split}
\max_{\theta_{\mathcal{P}},\theta_{\mathcal{W}},\mathbf{F}}  \mathbb{P}(\theta_{\mathcal{P}},\theta_{\mathcal{W}},\mathbf{Y,F}|\mathbf{X})\\
  \overset{\text{(1)}}{\thickapprox}\max_{\theta_{\mathcal{W}},\mathbf{\widetilde{F}}}\mathbb{P}(\mathbf{Y}|\mathbf{X,\widetilde{F}},\theta_{\mathcal{W}})\mathbb{P}(\theta_{\mathcal{W}})
 \overset{\text{(2)}}{\thickapprox}\max_{\theta_{\mathcal{W}},\mathbf{\widetilde{F}}}\mathbb{P}(\mathbf{Y}|\mathbf{\widetilde{F}},\theta_{\mathcal{W}})\mathbb{P}(\theta_{\mathcal{W}}),\\ \mathrm{s.t.}~\widetilde{\mathbf{F}}\in\arg\max_{\theta_{\mathcal{P}},\mathbf{F}}\mathbb{P}(\mathbf{F}|\mathbf{X},\theta_{\mathcal{P}})\mathbb{P}(\theta_{\mathcal{P}})
 \end{split}
\end{equation}
where we conduct an approximation imitating the strategy in \eqref{Equ:Classification_parametric} with a clear interpretation, \emph{i.e.}, find a MAP solution ${\widetilde{\mathbf{F}}}$ of a proper parametric feature model in the lower-level optimization to simultaneously maximize the upper level problem $\mathbb{P}(\mathbf{Y}|\mathbf{X,\widetilde{F}},\theta_{\mathcal{W}})$ so that the feature mapping will be consistent. In the following part, we will respectively design the upper level optimization corresponding to classifier construction and lower-level parametric feature model in the training phase.
\subsubsection{Parametric Classifier Model}  Let $\mathbf{y}_i\in\{0,1\}^C$ be a binary label vector of training sample $\mathbf{x}_i$, where it is the $c$-th column of identity matrix $\mathbf{I}\in\mathbb{R}^{C\times C}$ if $\mathbf{x}_i$ belongs to the $c$-th class. Focusing on the first type in \eqref{Equ:Classification_parametricposteriorrelaxition1}, a class-specific model will be learned for the samples from the same class. More specifically, we aim to learn $C$ set of independent model parameters $\{\theta_{\mathcal{W}}^c,\theta_{\mathcal{P}}^c\}_{c=1}^C$ to characterize the log-posterior distribution for $i$-th sample as:
 \begin{equation}\label{Equ:FirstType_DNAOL}
 \begin{split}
-\Pi_{c=1}^C\log \mathbb{P}(\mathbf{y}_i|\mathbf{x}_i,\theta_{\mathcal{W}}^c,\widetilde{\mathbf{f}}_{i}^c)=\sum_{c=1}^C\mathbf{y}_i(c)\ell(\mathbf{x}_i,\mathbf{W}^c\widetilde{\mathbf{f}}_{i}^c)\\
\mathrm{s.t.}~\widetilde{\mathbf{f}}_{i}^c\in\arg\max_{\mathbf{f},\theta_{\mathcal{P}}^c} \mathbb{P}(\mathbf{f}|\mathbf{x}_i,\theta_{\mathcal{P}}^c)
\end{split}
 \end{equation}
where $\mathbf{W}^c\in\mathbb{R}^{n\times p_c}$ denotes by the classifier parameters for $c$-th class, $\mathbf{\widetilde{f}}_i^c$ is the cosparse feature vector of $\mathbf{x}_i$ obtained with the $c$-th model and $\ell(\cdot,\cdot)$ is a suitable loss function measuring the fitting error. It is noted that \eqref{Equ:FirstType_DNAOL} will reach its lower bound if the loss function is vanishing as $\mathbf{x}_i=\mathbf{W}^c\widetilde{\mathbf{f}}_{i}^c$ and minimizing \eqref{Equ:FirstType_DNAOL} will encourage a least intra-class fitting error. Nevertheless, note also from the fact that as $\mathbf{y}_i(c)=0$ for samples belonging to the other class, such loss function in \eqref{Equ:FirstType_DNAOL} will make no contribution to enhancing the inter-class variances. We will address this problem later by imposing a constraint on $\theta_{\mathcal{P}}^c$. This type of framework will be denoted by Sep-DNAOL and it can be concluded that different model parameters are designed and operated independently no matter they are coherent or not. Considering the second type in \eqref{Equ:Classification_parametricposteriorrelaxition1} termed NonSep-DNAOL, we will exploit a more aggressive way by measuring the classification loss with a non-separable feature vector with the following resulted log-posterior distribution:
\begin{equation}\label{Equ:SecondType_DNAOL}
 \begin{split}
 \setlength{\abovedisplayskip}{1pt}
-\log \mathbb{P}(\mathbf{y}_i|\theta_{\mathcal{W}},\widetilde{\mathbf{f}}_{i})=\ell (\mathbf{y}_i, \mathbf{W}\widetilde{\mathbf{f}}_{i}),~
\mathrm{s.t.}~\widetilde{\mathbf{f}}_{i}\in\arg\max_{\mathbf{f},\theta_{\mathcal{P}}} \mathbb{P}(\mathbf{f}|\mathbf{x}_i,\theta_{\mathcal{P}})
\end{split}
\end{equation}
where $\mathbf{W}\in\mathbb{R}^{C\times p}$ is the parameters in the classifier. For both schemes of \eqref{Equ:FirstType_DNAOL} and \eqref{Equ:SecondType_DNAOL}, following the common strategy in previous DL algorithm \cite{aharon2006svd,mairal2010online}, we will also impose a unit $\ell_2$ norm constraint on each column of $\mathbf{W}$ or $\mathbf{W}^c$ to model $\mathbb{P}(\theta_{\mathcal{W}})$, \emph{i.e.}, $\|\mathbf{W}_{:,j}\|_2^2\leq 1$, which will be equivalent to imposing a weight decay regularization on the classifier parameters to prevent from over-fitting. In two schemes, the square $\ell_2$ loss function is considered for simplicity, namely $\ell(\mathbf{x,y})=\|\mathbf{x}-\mathbf{y}\|_2^2$ and it is worth noting that the above presented two types classifiers are both inherited from the previous DL algorithms without any new components being designed and involved.

\subsubsection{Feature Model}
\par After designing the upper-level optimization for parametric classifier model, we come to the most important issue in this framework, namely the lower-level MAP estimation of feature vectors and parameters $\theta_{\mathcal{P}}$ with a proper feature model. In general, it requires some complicated iterative methods between latent features and model parameters to pursuit the solutions, \emph{e.g.}, Expectation Maximization (EM) algorithm \cite{Bishop2006}. Fortunately, thanks to the proposed NACM, it has been declared in the previous section that this model can directly provide a closed form solution of the features to reach the lower bound of the posterior function with respect to $\theta_{\mathcal{P}}$. Therefore, the lower-level problem can be always simplified as the MAP estimation only with respect to $\theta_{\mathcal{P}}$ so that \eqref{Equ:Classification_parametricposteriorrelaxition1} will be reformulated as:
\begin{equation}\label{Equ:Classification_parametricposteriorrelaxition}\small
\begin{split}
\max_{\theta_{\mathcal{P}},\theta_{\mathcal{W}},\mathbf{F}}  \mathbb{P}(\theta_{\mathcal{P}},\theta_{\mathcal{W}},\mathbf{Y,F}|\mathbf{X})\\
  \overset{\text{(1)}}{\thickapprox}\max_{\theta_{\mathcal{W}},\theta_{\mathcal{P}},\mathbf{\widetilde{F}}}\sum_{c=1}^C\sum_{i=1}^N\mathbf{y}_i(c)\ell(\mathbf{x}_i,\mathbf{W}^c\widetilde{\mathbf{f}}_{i}^c)+\frac{\alpha}{2}\|\mathbf{A}^c\mathbf{\overline{X}}^c\|_\mathrm{F}^2+\frac{\tau}{2}\|\mathbf{A}^c\|_\mathrm{F}^2,\\
   \mathrm{s.t.}~\mathbf{\widetilde{F}}^c=\mathcal{S}_{\mathbf{\Lambda}^c}(\mathbf{A}^c\mathbf{X}),~\|\mathbf{W}_{:,j}^c\|_{2}^2\leq 1\\
 \overset{\text{(2)}}{\thickapprox}\max_{\theta_{\mathcal{W}},\theta_{\mathcal{P}},\mathbf{\widetilde{F}}}\sum_{i=1}^N\ell (\mathbf{y}_i, \mathbf{W}^c\widetilde{\mathbf{f}}_{i}^c)+\tau\|\mathbf{A}\|_\mathrm{F}^2,\\
 \mathrm{s.t.}~\mathbf{\widetilde{F}}=\mathcal{S}_{\mathbf{\Lambda}}(\mathbf{AX}),~\|\mathbf{W}_{:,j}\|_{2}^2\leq 1
 \end{split}
\end{equation}
where we impose Gaussian distributed prior on each parameter in $\mathbf{A}$, yielding an essentially weight decay regularization with the hyper-parameter $\tau$. The term $\alpha\|\mathbf{A}^c\mathbf{\overline{X}}^c\|_\mathrm{F}^2$ is involved to encourage the inter-class dissimilarity to remedy the deficiency of the classifier model, where $\mathbf{\overline{X}}^c$ contains the samples excluding class $c$ and $\alpha$ is another hyper-parameter. More substantial, this term will enforce the samples to be unfitted with an inter-class feature model.
\par The remaining issue will focus on the explicit expression of the feature selector $\mathcal{S}_{\mathbf{\Lambda}}$. Inspired from the soft-threshold operator corresponding to the proximity operator of the $\ell_1$ norm, $\mathcal{S}_{\mathbf{\Lambda}}$ will be designed as the following parametric scaled-threshold function in this paper\footnote{It is worth noting that other choices will be also available and encouraged.}.
\begin{equation}\label{Equ:Selector}
\begin{split}
  \mathcal{S}_{\mathbf{\Lambda}}(\mathbf{Ax})&=\mathbf{\Lambda}_1\mathrm{sgn}(\mathbf{\widetilde{A}x})\max(|\mathbf{\widetilde{A}x}|-\mathbf{\Lambda_2},{\mathbf{0}}),\\
  &=\mathbf{\Lambda}\mathrm{sgn}(\mathbf{Ax})\max(|\mathbf{Ax}|-\mathbf{1},\mathbf{0})
\end{split}
\end{equation}
where $\mathbf{\Lambda}\in\mathbb{R}^{p\times p}$ is a diagonal parametric matrix controlling the scale of each feature response and the threshold values for each component simultaneously. We can easily observe that the soft-threshold function is a special instance of \eqref{Equ:Selector} in the case of $\mathbf{\Lambda}=\mathbf{I}$ with threshold parameter vector as $\mathbf{1}$. It follows that the $j$-th feature response in NACM will be independently computed as:
\begin{equation}\label{Equ:Featurejthresponse}
  \mathbf{f}(j)=\lambda_j \mathrm{sgn}(\mathbf{A}_{j,:}\mathbf{x}) \max(|\mathbf{A}_{j,:}\mathbf{x}|-1)
\end{equation}
where $\lambda_j$ is the $j$-th entry in the diagonal of $\mathbf{\Lambda}$. When $|\mathbf{A}_{j,:}\mathbf{x}|<1$, the corresponding feature response will vanish so that the robustness of the sparse pattern in the feature vector will be promoted than conventional ACM. Compared with \eqref{Equ:Solu_PCR}, we can observe that $\mathbf{\Lambda}$ in \eqref{Equ:Selector} will not only perform as the scale parameters for each feature response, but also the regularization parameter in \eqref{Equ:Solu_PCR}, which further enables us to learn the regularization parameter according to our task. In this paper, since we have no preference for the selected features, we will impose on each $\lambda_j$ a constraint that each of them will be equal and drawn from a uniform distribution, namely $\lambda_j=\lambda_i\sim \mathcal{U}(0,1),~\forall j\neq i$, yielding a scalar parameter to simplify our model.
\subsection{Classification in the Testing Phase}
\par In the previous part, we have completed the framework of two types of DNAOL in the training phase, which will result in the MAP solution of parameters $\theta_{\mathcal{W}}^*$ and $\theta_{\mathcal{P}}^*$. In this subsection, we will present the testing phase for classifying a query sample by considering the following MAP estimation.
\begin{equation}\label{Equ:Testingphase}
\begin{split}
  \max_{\mathbf{\widehat{y}},\mathbf{\widehat{f}}} \mathbb{P}(\mathbf{\widehat{y}},\mathbf{\widehat{f}}|\widehat{\mathbf{x}},\theta_{\mathcal{P}}^*,\theta_{\mathcal{W}}^*)\overset{\text{(1)}}{\thickapprox} \max_{\widehat{\mathbf{y}}} \mathbb{P}(\widehat{\mathbf{y}}|\widehat{\mathbf{f}}^*,\widehat{\mathbf{x}},\theta_{\mathcal{W}}^*)\\
  \overset{\text{(2)}}{\thickapprox}\max_{\widehat{\mathbf{y}}}\mathbb{P}(\widehat{\mathbf{y}}|\widehat{\mathbf{f}}^*,\theta_{\mathcal{W}}^*)~
  \mathrm{s.t.}~\widehat{\mathbf{f}}^*=\arg\max \mathbb{P}(\mathbf{f}|\widehat{\mathbf{x}},\theta_{\mathcal{P}}^*)\\
  \end{split}
\end{equation}
For Sep-DNAOL, a set of parameters for each class will be denoted by $(\theta_{\mathcal{W}}^c)^*=(\mathbf{W}^c)^*$ and $(\theta_{\mathcal{P}}^c)^*=\{(\mathbf{A}^c)^*,(\mathbf{\Lambda}^c)^*\}$. Then the label vector $\widehat{\mathbf{y}}$ of a query sample $\widehat{\mathbf{x}}$ will consequently be inferred as:
\begin{equation}\label{Equ:Classfication1}
  \widehat{\mathbf{y}}=\arg\min_c\|\mathbf{\widehat{x}}-(\mathbf{W}^c)^*\mathcal{S}_{(\mathbf{\Lambda}^c)^*}((\mathbf{A}^c)^* \widehat{\mathbf{x}}) \|_2^2.
\end{equation}
For NonSep-DNAOL, we will directly check the classification loss via:
\begin{equation}\label{Equ:LossRegression2}
  \widehat{\mathbf{y}}=\arg\max_c \mathbf{W}_{c,:}^*\mathcal{S}_{\mathbf{\Lambda}^*}(\mathbf{A}^* \widehat{\mathbf{x}})
\end{equation}
It can be observed from \eqref{Equ:Classfication1} and \eqref{Equ:LossRegression2} that the feature mapping in the testing phase for both classification schemes are consistent with those in the corresponding training phase and the  computation cost will be cheap due to a series of simple forward operations with the closed form, namely matrix-vector multiplication and logical comparison.
\subsection{Optimization Scheme}
\par In this subsection, we will derive a detailed optimization scheme for two schemes or DNAOL frameworks in \eqref{Equ:Classification_parametricposteriorrelaxition}.  Considering the problem, both schemes will actually consist of two coupled procedures, \emph{i.e.}, parametric NACM learning by optimizing $\mathbf{A}$ and $\mathcal{S}_{\mathbf{\Lambda}}$ and classifier learning by updating $\theta_{\mathcal{W}}$. Following common strategies in the previous DL algorithm, we will achieve this goal by alternatively dealing with two modules, \emph{i.e.}, optimizing the classifier with the resulted features to minimize the classification loss, and then updating the feature model to produce the MAP solutions of $\mathbf{F}$ and $\mathbf{A}$ with respect to the current $\theta_{\mathcal{W}}$. It follows that the overall problem will be non-convex so that only a local optimal could be reached depending on the initialization. In our framework, we initialize $\mathbf{W}$ with the standard Gaussian random matrix and then project each column into the unit $\ell_2$ ball to make it feasible. $\mathbf{A}$ will be also initialized as the Gaussian random matrix whose variance will be determined by the training databases. The nonlinear parameters in $\mathbf{\Lambda}$ will be initialised with the values drawn from the uniform distribution $\mathcal{U}(0,1)$.
\par Given a fixed $\mathbf{W}$, the optimization problem with respect to NACM will be roughly summarized as:
\begin{equation}\label{Opt:Firstphase}
  \min_{\mathbf{A},\mathbf{\Lambda},\mathbf{F}} \ell(\mathbf{F})+\Psi(\mathbf{A}),~\mathrm{s.t.}~\mathcal{S}_{\mathbf{\Lambda}}(\mathbf{AX})-\mathbf{F}=\mathbf{0},
\end{equation}
where $\ell(\mathbf{F})$ is short for the loss functions with respect to $\mathbf{F}$ in two schemes here and after in the case of $\mathbf{W}$ is fixed, vice versa and $\Psi(\mathbf{A})$ is the regularizations on $\mathbf{A}$. In order to solve this problem, the general alternating direction method of multipliers (ADMM) will be exploited, which will be beneficial to the large scaled problem with distributed computing scheme \cite{Boyd2011}. In order to make it easier to handle, an auxiliary variable $\mathbf{Z}$ is introduced for $\mathbf{AX}$ and the final augmented Lagrangian function is given by:
\begin{equation}\label{Equ:Lagrangian}
\begin{split}
  \mathcal{L}_{\rho}= \ell(\mathbf{F})+\Psi(\mathbf{A})+<\mathbf{U}_1,\mathbf{Z-AX}>\\
  +<\mathbf{U}_2,\mathcal{S}_{\lambda}(\mathbf{Z})-\mathbf{F}>+\frac{\rho}{2}\|\mathbf{Z-AX}\|_\mathrm{F}^2+\frac{\rho}{2}\|\mathcal{S}_{\lambda}(\mathbf{Z})-\mathbf{F}\|_\mathrm{F}^2,
  \end{split}
\end{equation}
where $<\mathbf{A},\mathbf{B}>=\mathrm{Tr}(\mathbf{A}^\mathrm{T}\mathbf{B})$ stands for the inner product operation between $\mathbf{A}$ and $\mathbf{B}$, $\mathbf{U}_1$ and $\mathbf{U}_2$ are two dual lagrangian multipliers and $\rho>0$ is a penalty parameter. Then the corresponding dual problem with respect to $\mathbf{U}_1$ and $\mathbf{U}_2$ are given by:
\begin{equation}\label{Opt:Dualproblem}
  \max_{\mathbf{U}_1,\mathbf{U}_2} \inf_{\mathbf{F},\mathbf{A},\mathbf{Z},\mathbf{\Lambda}} \mathcal{L}_{\rho}(\mathbf{F,A},\mathbf{Z},\mathbf{\Lambda},\mathbf{U}_1,\mathbf{U}_2)
\end{equation}
Totally speaking, the optimization process with ADMM will be conducted according to the following scheme:
\begin{equation}\label{Opt:TotalScheme}\small
  \begin{cases}
  \mathbf{F}^{(k+1)}=\arg\min_{\mathbf{F}}\ell(\mathbf{F})+\frac{\rho}{2}\|\mathcal{S}_{\mathbf{\Lambda}^{(k)}}(\mathbf{Z}^{(k)})-\mathbf{F}+\frac{\mathbf{U}_2^{(k)}}{\rho}\|_\mathrm{F}^2,\\
   \mathbf{Z}^{(k+1)}=\arg\min_{\mathbf{Z}}\|\mathcal{S}_{\mathbf{\Lambda}^{(k)}}(\mathbf{Z})-\widetilde{\mathbf{U}}_2^{(k)}\|_\mathrm{F}^2+\|\mathbf{Z}-\widetilde{\mathbf{U}}_1^{(k)}\|_\mathrm{F}^2,\\
   \mathbf{\Lambda}^{(k+1)}=\arg\min_{\mathbf{\Lambda}}\|\mathcal{S}_{\mathbf{\Lambda}}(\mathbf{Z}^{(k+1)})-\widetilde{\mathbf{U}}_2^{(k)}\|_\mathrm{F}^2\\
    \mathbf{A}^{(k+1)}=\arg\min_{\mathbf{A}}\Psi(\mathbf{A})+\frac{\rho}{2}\|\mathbf{Z}^{(k+1)}-\mathbf{AX}+\frac{\mathbf{U}_1^{(k)}}{\rho}\|_\mathrm{F}^2,\\
  \mathbf{U}_1^{(k+1)}=\mathbf{U}_1^{(k)}+\rho(\mathbf{Z}^{(k+1)}-\mathbf{A}^{(k+1)}\mathbf{X}),\\
  \mathbf{U}_2^{(k+1)}=\mathbf{U}_2^{(k)}+\rho(\mathcal{S}_{\mathbf{\Lambda}^{(k+1)}}(\mathbf{Z}^{(k+1)})-\mathbf{F}^{(k+1)}),\\
 \end{cases}
\end{equation}
where $\widetilde{\mathbf{U}}_1^{(k)}=\mathbf{A}^{(k)}\mathbf{X}-\frac{\mathbf{U}_1^{(k)}}{\rho}$, $\widetilde{\mathbf{U}}_2^{(k)}=\mathbf{F}^{(k+1)}-\frac{\mathbf{U}_2^{(k)}}{\rho}$ and the superscript $(k)$ or $(k+1)$ denotes the iterations in ADMM. We will respectively derive the detailed optimization schemes for each primal variable in the following, where each of them will admit an simple closed form solution without need any iteration process.

\begin{itemize}
  \item \emph{Update feature vectors} $\mathbf{F}$
  \par According to the optimality condition, the solution $\mathbf{F}^{(k+1)}$ should satisfy:
  \begin{equation}\label{Opt:F}
  \begin{split}
    \mathbf{0}&\in \frac{\partial \ell(\mathbf{F}^{(k+1)})}{\partial\mathbf{F}}-\rho\left(\mathcal{S}_{\mathbf{\Lambda}^{(k)}}(\mathbf{Z}^{(k)})-\mathbf{F}^{(k+1)}+\frac{\mathbf{U}_2^{(k)}}{\rho}\right).
    \end{split}
  \end{equation}
 Since $\ell(\mathbf{F})$ in the both classification schemes are quadratic functions, the optimal $\mathbf{F}^{(k+1)}$ can be straightforwardly computed by solving the linear Eq. \eqref{Opt:F}.
  \item \emph{Update auxiliary variable} $\mathbf{Z}$
  \par The optimization for this variable will be more sophisticated than the previous ones due to the nonlinear function. Nonethless, considering the particular form of our designed selector, we will next show that a closed form solution can be obtained in the case of such a piecewise linear function $\mathcal{S}_{\mathbf{\Lambda}}$. Examining the subproblem with respect to $\mathbf{Z}$, we can observe that it is separable with respect to each entry in $\mathbf{Z}$ so that the problem for $\mathbf{Z}_{j,i}$ will be given by:
  \begin{equation}\label{Opt_Z}
    \min_{\mathbf{Z}_{j,i}} (\mathcal{S}_{\lambda_j}(\mathbf{Z}_{j,i})-(\widetilde{\mathbf{U}}_2)^{(k)}_{j,i})^2+(\mathbf{Z}_{j,i}-(\widetilde{\mathbf{U}}_1)_{j,i}^{(i)})^2
  \end{equation}
  It follows that \eqref{Opt_Z} will become a piecewise quadratic function whose local minimizers can be readily computed with the closed form. After each local minimal is obtained, the global optimal $\mathbf{Z}_{j,i}$ will be the one among those local solutions via logical comparison.
    \item \emph{Update the nonlinear feature selector }$\mathcal{S}_{\mathbf{\Lambda}}$\\
     The subproblem with respect to $\mathcal{S}_{\mathbf{\Lambda}}$ is summarized as following:
  \begin{equation}\label{Opt_lambda}
  \begin{split}
   \min_\mathbf{\Lambda} \|\mathcal{S}_{\mathbf{\Lambda}}(\mathbf{Z}^{(k+1)})-\widetilde{\mathbf{U}}_2^{(k)}\|_\mathrm{F}^2
   \end{split}
  \end{equation}
  Considering the above optimization, it is also a quadratic function with respect to the scalar $\lambda$. Therefore, the solution is obtained via setting its derivative as zero and solving the linear equation.
  \item \emph{Update the linear feature extractor }$\mathbf{A}$
  \par Analogously to the optimization process for $\mathbf{F}$, the closed form solution of $\mathbf{A}^{(k+1)}$ can be also obtained by solving the following equation when the regularizer $\Psi(\mathbf{A})$ is a quadratic function.
    \begin{equation}\label{Opt:A}\small
    \begin{split}
    \mathbf{0}&\in\partial\Psi(\mathbf{A}^{(k+1)})-\rho\left(\mathbf{Z}^{(k+1)}-\mathbf{A}^{(k+1)}\mathbf{X}+\frac{\mathbf{U}_1^{(k)}}{\rho}\right)\mathbf{X}^\mathrm{T}\\
    &=\partial\Psi(\mathbf{A}^{(k+1)})-\mathbf{\Lambda}_1^{(k+1)}\mathbf{X}^\mathrm{T}.
    \end{split}
  \end{equation}
    \par ADMM converges when the primal and dual residuals are both below a threshold or the total iterations exceed its maximum.
  \end{itemize}
\par After optimizing the NACM, we will approximately obtain the MAP solution of $\mathbf{F}$ and $\theta_{\mathcal{P}}$, and then we are ready to optimize the classifier as:
  \begin{equation}\label{Opt:W}
    \min_{\mathbf{W}} \ell(\mathbf{W})~\mathrm{s.t.}~\|\mathbf{W}_{j,:}\|_2^2\leq 1
  \end{equation}
The solution to this constraint convex optimization can be simply computed via projected gradient algorithm. Otherwise, we can also exploit the ADMM schemes by introducing another variable as \cite{Gu2014} does and we will not discuss the details for simplicity. The overall algorithm converges when the relative variation of loss function between two adjacent iterations is below a predefined threshold $\epsilon$ and it is summarised in Algorithm \ref{ALG:algorithm1}. In practical situation, we can exploit the warm start strategy for ADMM to search a good initializations, namely optimizing each primal variable only without Lagrangian multipliers updating.

\begin{algorithm}[b]\caption{DNAOL}\label{ALG:algorithm1}

\KwIn{Training set $(\mathbf{X},\mathbf{Y})$, regularization parameters $\alpha$, $\tau$, max iterations $T$, penalty parameter $\rho$, residual $\epsilon$.}
\KwOut{$\mathbf{A},~\mathbf{W},~\mathbf{\Lambda}=\{\lambda\}$}
\textbf{Initialize:}
{$\mathbf{W}\sim\mathcal{N}(\mathbf{0},\mathbf{I})$, $\lambda\sim \mathcal{U}(0,1)$, $\mathbf{A}\sim\mathcal{N}(\mathbf{0},\sigma^2 \mathbf{I})$

}
\textbf{Main Loop:}~
\While{ not convergence}
{
    Update $\mathbf{A}$, $\mathcal{S}_{\mathbf{\Lambda}}$ and $\mathbf{F}$ with scheme \eqref{Opt:TotalScheme}\;
    Update $\mathbf{W}$ by solving \eqref{Opt:W}\;
}
\end{algorithm}

\subsection{Framework Comparison}

\begin{figure}
  \centering
  \includegraphics[width=0.5\textwidth]{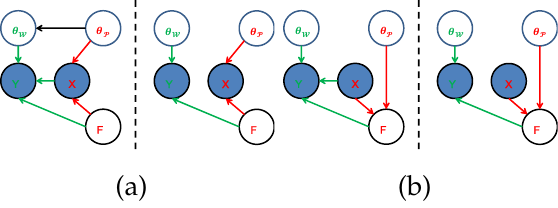}\\
  \caption{Probabilistic graphical representations for two types of classification frameworks with (a). generative version with regularized linear synthesis model and (b). discriminative version with nonlinear analysis cosparse model. Green and red arrays will stand for the procedures in the classifier and feature model, respectively.}\label{Fig_Graphic}
\end{figure}

In this subsection, we will make a comparison with two typical counterpart DL frameworks and AOL classification frameworks to highlight the novelty and superiorities of DNAOL. Before we start, probabilistic graphic representations for generative and discriminative models with two classification implementations are shown in Figs. \ref{Fig_Graphic} to better illustrate the differences.
\par Considering Sep-DNAOL, we essentially learn a class-specific model for each class independently, resulting in $C$ set of model parameters $\{\theta_{\mathcal{W}}^c,\theta_{\mathcal{P}}^c\}_{c=1}^C$, where we only impose a discriminative constraint on $\mathbf{A}^c$ to increase the inter-class dissimilarity. It will make us recall a similar DL framework called dictionary learning with structured incoherence (DLSI) which will also attempt to learn $C$ distinct synthesis models (dictionaries) \cite{ramirez2010classification}. To enhance the discrimination, the authors introduced a structured regularization on each pair of dictionary to increase their mutual incoherences. From this perspective, DLSI should be regarded as the twin framework with Sep-DNAOL, however, we will see from the following empirical results that Sep-DNAOL will significantly outperform DLSI in terms of accuracies and execution times. Considering the NonSep-DNAOL, it is much more similar to discriminative K-SVD (D-KSVD) \cite{zhang2010discriminative}, where only a label regression term is involved in both frameworks without extra supervised strategy. Note also from their following experimental results that NonSep-DNAOL will achieve much better performances. These two instructive examples will evidently demonstrate the superiorities of the proposed NACM over the regularized linear synthesis model in general classification task. In addition to these two DL frameworks, we will further observe that the DNAOL can achieve the better or competitive results than other state-of-the-art generative DL frameworks involving more discriminative strategies.
\par Next, let us review the framework of DPL \cite{Gu2014} with the purpose of learning a linear analysis operator to approximate the features of a synthesis dictionary. This framework is also a discriminative one focusing on the posterior distribution, but its motivation will be distinct from our Sep-DNAOL. Moreover, Sep-DNAOL enables us to produce a MAP solution of the features with task-adapted regularization in a latent way while features in DPL are the intermediate latent variables obtained with a maximizing likelihood estimation (MLE) so that the inconsistency issue will still appear in their framework. We will also compare NonSep-DNAOL with fast soft-thresholding based dictionary learning (ST-DL) \cite{fawzi2014dictionary}. Intuitively speaking, apart from their distinct motivations, ST-DL can be also viewed as a special instance of NonSep-DNAOL, but the classification loss functions in two frameworks are different. More specifically, ST-DL exploits the hinge loss for binary classification while we leverage the simple ridge regression function for multi-class classification. More importantly, ST-DL uses a fixed non-parametric soft-thresholding operator as the nonlinear mapping without learning while we focus on learning a task adapted parametric scaling soft-threshold function according to some training samples. Therefore, generality and adaptability of NonSep-DNAOL can be fundamentally promoted than ST-DL.
\section{Experiments}\label{Sec:Experiment}
In this section, we will experimentally evaluate the proposed DNAOL framework for different perspectives. We will first examine the convergence and computational complexities of the proposed framework. Next, the effects of different parameters are investigated . Finally, several state-of-the-art DL-SSM frameworks are compared to demonstrate the superiorities of DNAOL. All the experiments are performed on a desktop PC with i5 Intel CPU and 8GB memory for ten times independently and the average results will be reported. Without loss of the generality, the following experiments for analysis will be conducted on Extended Yale B database.
\subsection{Convergence Analysis}
\par In this subsection, we will analyze the convergence of the optimization scheme for DNAOL. It has been declared that the overall optimization scheme will converge to the local optimal solutions due to the non-convexity of the problem, therefore the convergence curves will be empirically plotted in Fig. 2(a), implying the increase of the log posterior probabilities. Considering the nonlinear analysis operator learning phase, the corresponding subproblem \eqref{Opt:Firstphase} is a nearly canonical form for ADMM whose convergence has been extensively investigated \cite{Boyd2011}. Specifically, it indicates that the primal and dual residuals will converge to zero to ensure the optimality conditions for each subproblem in ADMM scheme with $\mathcal{O}(1/k)$ rate for convex problem. We empirically plot the primal and dual residual curves in the first main iteration for Sep-DNAOL and NonSep-DNAOL in Figs. 2(b) and 2(c), respectively. We can see that the primal residuals (PR1,PR2) for both schemes will converge fast to zero while the dual residuals (DR1, DR2) will only reach stability. This phenomenon is probably caused by alternatively solving the nonlinear subproblems with respect to $\mathbf{Z}$ and $\mathbf{\Lambda}$. Another important issue should be discussed here that a larger value of $\rho$ will impose a larger penalty on the primal residuals to ensure the constraint conditions while a smaller one will help to reduce the dual residuals to reach a closer optimal for the dual problem. In our experiments, $\rho$ is fixed as $1$ for simplicity. For the subproblem of classifier learning \eqref{Opt:W}, it is a convex optimization which will converge to the global minimizer.

\begin{figure*}
  \centering
  \includegraphics[width=0.8\textwidth]{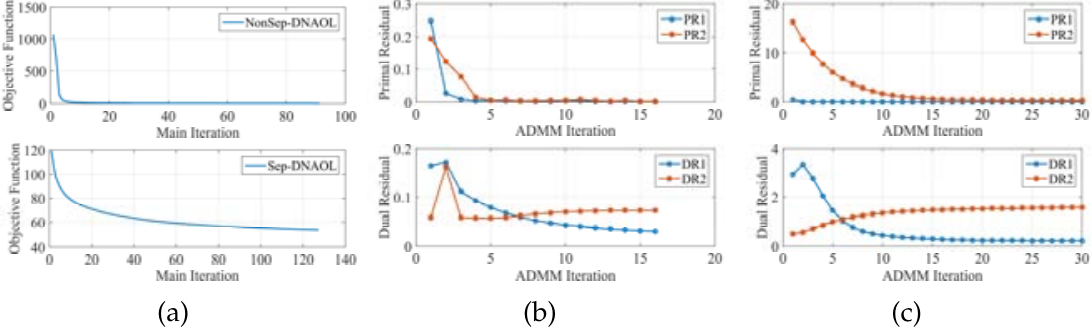}\\
  \caption{Convergence curves. (a). Objective function values for Sep-DNAOL and NonSep-DNAOL. (b). Primal and dual residuals for Sep-DNAOL. (c). Primal and dual residuals for NonSep-DNAOL.}\label{Fig_Convergence}
\end{figure*}

\subsection{Computational Complexity Analysis}
As one of the motivations and superiorities of DNAOL, the computational complexities in the training and testing phases should be examined, respectively. Reviewing the training phase in the first place, two classification schemes have been developed, namely Sep-DNAOL and NonSep-DNAOL. In Sep-DNAOL, we target at learning a set of completely separated parameters independently. On the contrary, NonSep-DNAOL will learn the common parameters for all classes so that its complexities will rely on all training samples. From this perspective, NonSep-DNAOL will generally more time consuming than Sep-DNAOL. Concerning each step of the optimization, the computational burden in nonlinear operator learning phase will mainly result from optimizing $\mathbf{F}$, $\mathbf{Z}$, $\lambda$ and $\mathbf{A}$. Specifically, updating $\mathbf{F}$ will rely on solving a linear equation in which we have to compute the inverse of $(\mathbf{W}^\mathrm{T}\mathbf{W}+\rho \mathbf{I})$. This computational complexity will be $\mathcal{O}(p_c^3)$ for Sep-DNAOL and $\mathcal{O}(p^3)$ for NonSep-DNAOL, respectively. As $p_c\ll p$, computing $\mathbf{F}$ for NonSep-DNAOL will have a much more expensive cost and we can exploit the conjugate gradient method to iteratively pursuit the solution, if necessary. Updating $\mathbf{Z}$ is a fully separated problem for both schemes, where we only have to optimize each component of $\mathbf{Z}$ by solving a scalar piecewise quadratic function and then making if-then comparisons. The complexity of updating $\mathbf{\Lambda}$ will mostly depend on the size of $\mathbf{Z}$. Updating $\mathbf{A}$ is also solving a linear equation, which will need to compute the inverse of $(\tau\mathbf{I}+(\rho-\alpha) \mathbf{X}^c({\mathbf{X}}^c)^\mathrm{T}+\alpha \mathbf{X}\mathbf{X}^\mathrm{T})$ or $(\tau\mathbf{I}+\rho \mathbf{X}\mathbf{X}^\mathrm{T})$ for Sep-DNAOL and NonSep-DNAOL, respectively. For both schemes, the complexity will be $\mathcal{O}(\max (n^3,n^2N))$ depending on the size of the input training samples. Fortunately, the required inversion will keep unchanged during the training process for both schemes so that we can precompute and store them in advance to accelerate the learning procedures. However, when this trick is exploited, the burden of storage will increase. We empirically observed that if the training samples are collected from numerous subjects in a relative high dimensional input domain, the time cost will contrarily increase in a limited memory environment. In the testing phase, both of the schemes will have much fewer complexities than conventional frameworks of DL-SSM. Specifically, we only need to exploit \eqref{Equ:Classfication1} or \eqref{Equ:LossRegression2}, in which almost $\mathcal{O}(n^2p_c)$ and $\mathcal{O}(pn)$ complexities are required for classifying a query sample. Therefore, NonSep-DNAOL will spend less time than Sep-DNAOL in the testing phase.
\par We will now empirically evaluate the influence of $p_c$ and $p$ by varying the equivalent feature dimension from $10\times 38$ to $100\times 38$, namely $\sum_c p_c$ for Sep-DNAOL and $p$ for NonSep-DNAOL and the results are plotted in following Figs. \ref{Fig:Computationalcomplexity}. As shown in Fig. \ref{Fig:Subdicnum_EYaleB_Time}, Sep-DNAOL indeed requires much less time cost in the training phase than Non-Sep-DNAOL, where the time cost in the case of 3800 equivalent feature dimension is still much fewer than 100s. It is worth noting that such time cost for Sep-DNAOL is able to be further reduced if the parallel computing is adopted. For the testing phase, both of the schemes will efficiently classify a query within $10^{-3}s$, but NonSep-DNAOL will spend less time than Sep-DNAOL, which has been analysed above.

 \begin{figure}
  \centering
  \subfigure[]{\includegraphics[width=0.24\textwidth]{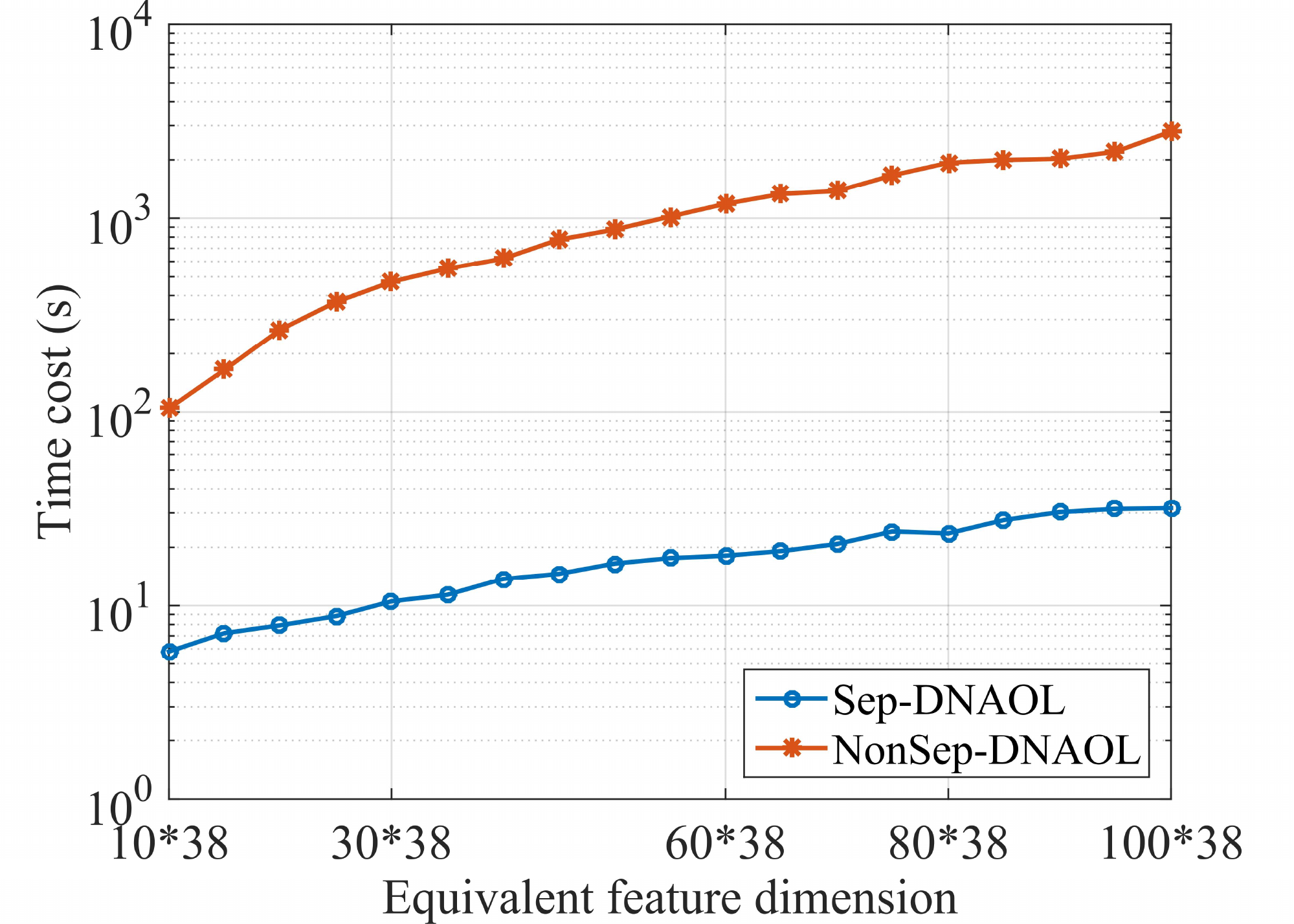}\label{Fig:Subdicnum_EYaleB_Time}}
  \subfigure[]{\includegraphics[width=0.24\textwidth]{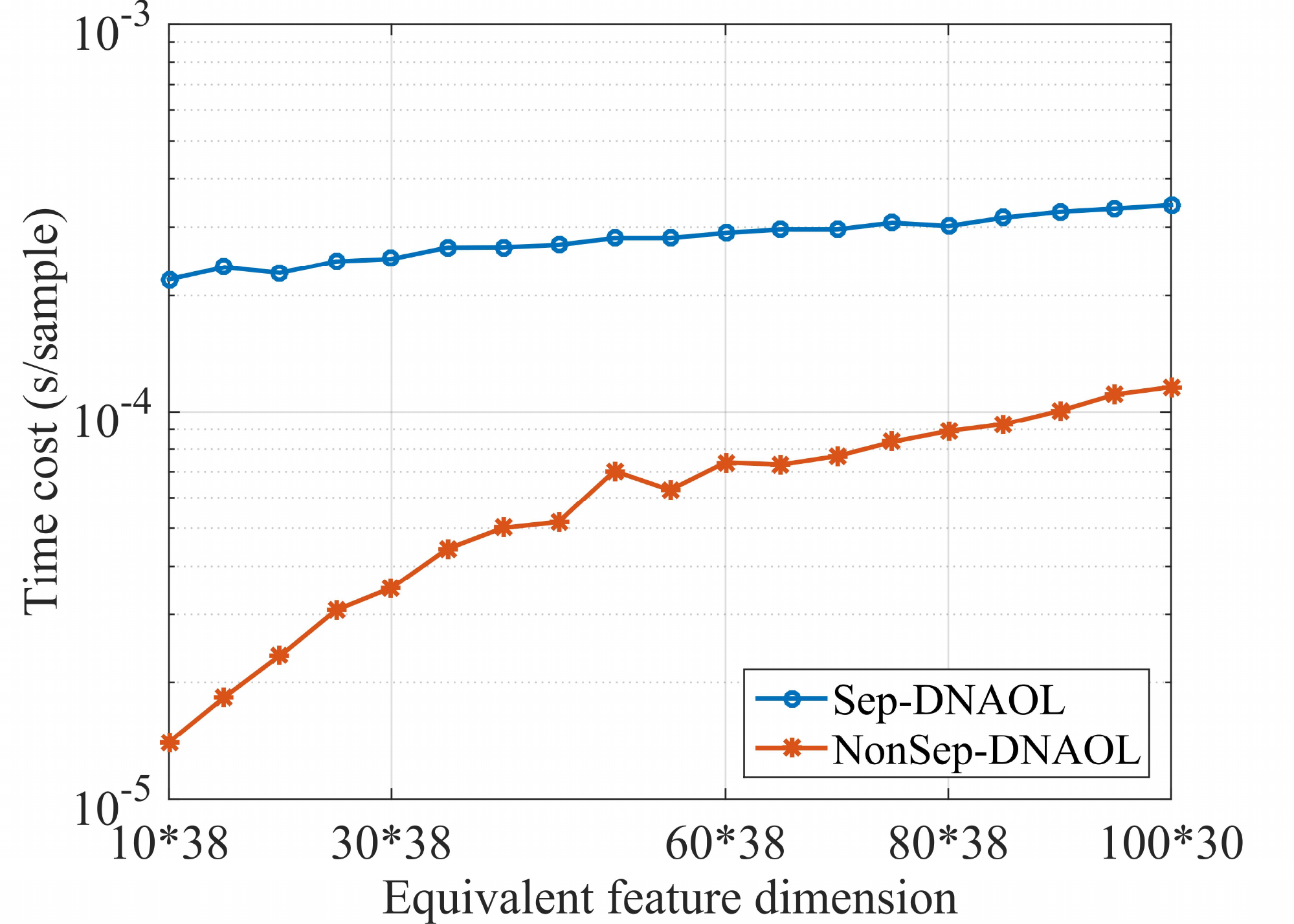}\label{Fig:Subdicnum_EYaleB_TestTime}}
  \caption{Time cost with different equivalent feature dimensions in (a) training  and  (b) testing phases, respectively.}\label{Fig:Computationalcomplexity}
\end{figure}

\subsection{Regularization Parameters}
The part will evaluate the effect of parameters in DNAOL, namely regularization parameters $\alpha$, $\tau$ and variance $\sigma$ for the initial $\mathbf{A}$. We firstly fix $\sigma=1$ and examine the effect of the regularization parameters, then with the local optimal $\alpha$ and $\tau$, we will further evaluate the influence of the initial variance.
\par
\begin{figure}
  \centering
   \subfigure[]{\includegraphics[width=0.23\textwidth]{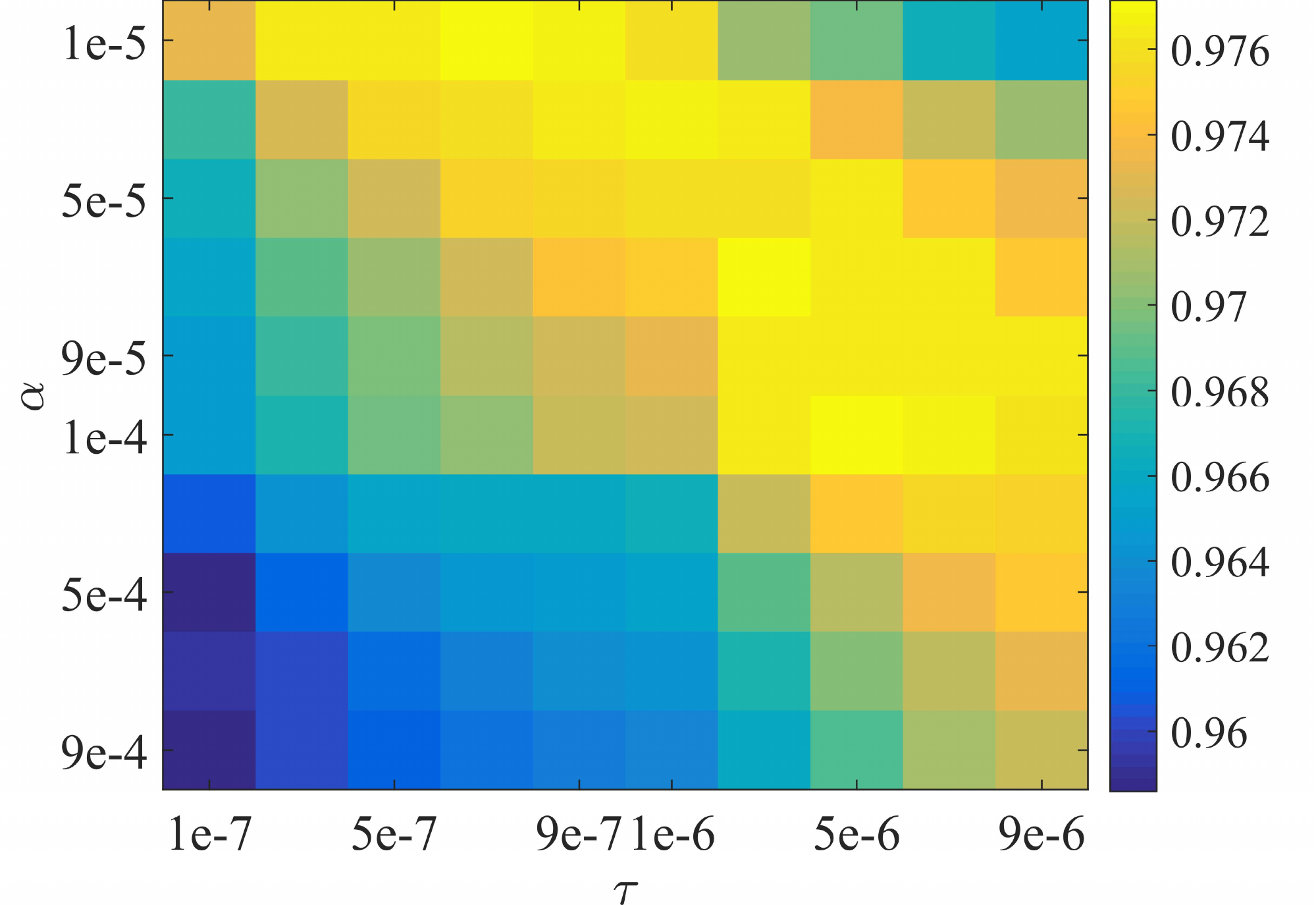}\label{Fig:Para_taualphaSepEYaleB}}
  \subfigure[]{\includegraphics[width=0.23\textwidth]{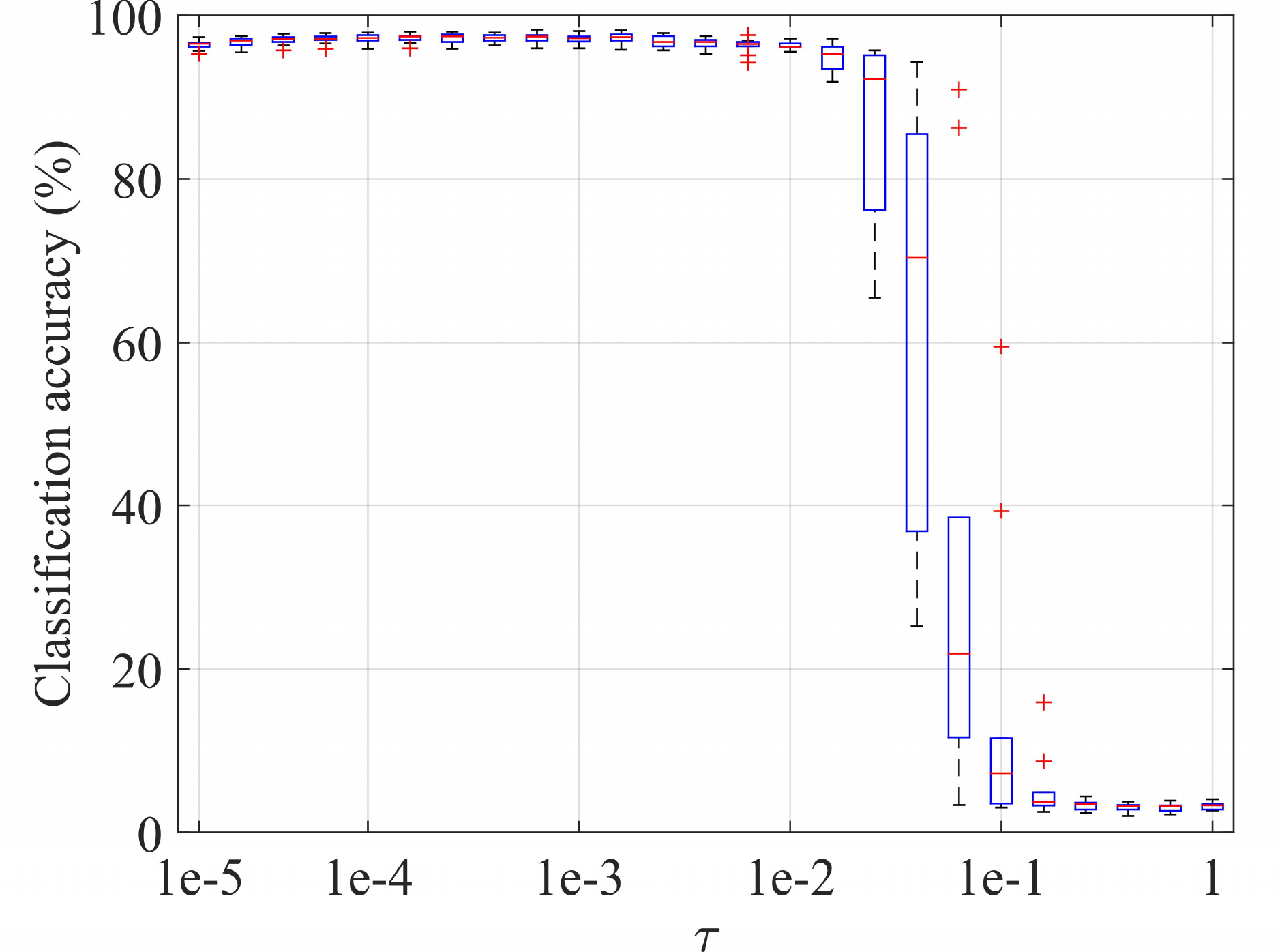}\label{Fig:Para_tauNonEYaleBBox}}
  \caption{Effects of hyper-parameters for two schemes on the classification accuracy. (a). the average classification performances of Sep-DNAOL. (b). box plot of the performances of NonSep-DNAOL.}\label{Fig:Paratau}
\end{figure}

 For Sep-DNAOL, we will exploit the grid search to tune $\alpha$ and $\tau$ in cross validation and the effects with respect to the classification accuracy are plotted in Fig. \ref{Fig:Para_taualphaSepEYaleB}. It can be seen from \eqref{Equ:Classification_parametricposteriorrelaxition}, the essential influence of $\alpha$ and $\tau$ for Sep-DNAOL will rely on their ratio. It can be observed from Fig. \ref{Fig:Para_taualphaSepEYaleB}, the values across each diagonal are approximately the same. Theoretically speaking, a larger value of ratio will help to enhance the discrimination of $\mathbf{A}_c$. Accordingly, we can conclude from the following figure that with the increase of this ratio, the accuracies are raising and reaching the peak regions (yellow region). However, too large a ratio will contrarily disturb the loss function so that the performance will be degraded gradually. For NonSep-DNAOL, we use the one dimensional search to determining the regularized path of $\tau$ and the results are plotted in Figs. \ref{Fig:Para_tauNonEYaleBBox}.As shown in the figure, the classification accuracies will be stable when $\tau<1e^{-2}$, where the performances gradually raise to the top as the increase of $\tau$.
\par
\begin{figure}
  \centering
   \subfigure[]{\includegraphics[width=0.23\textwidth]{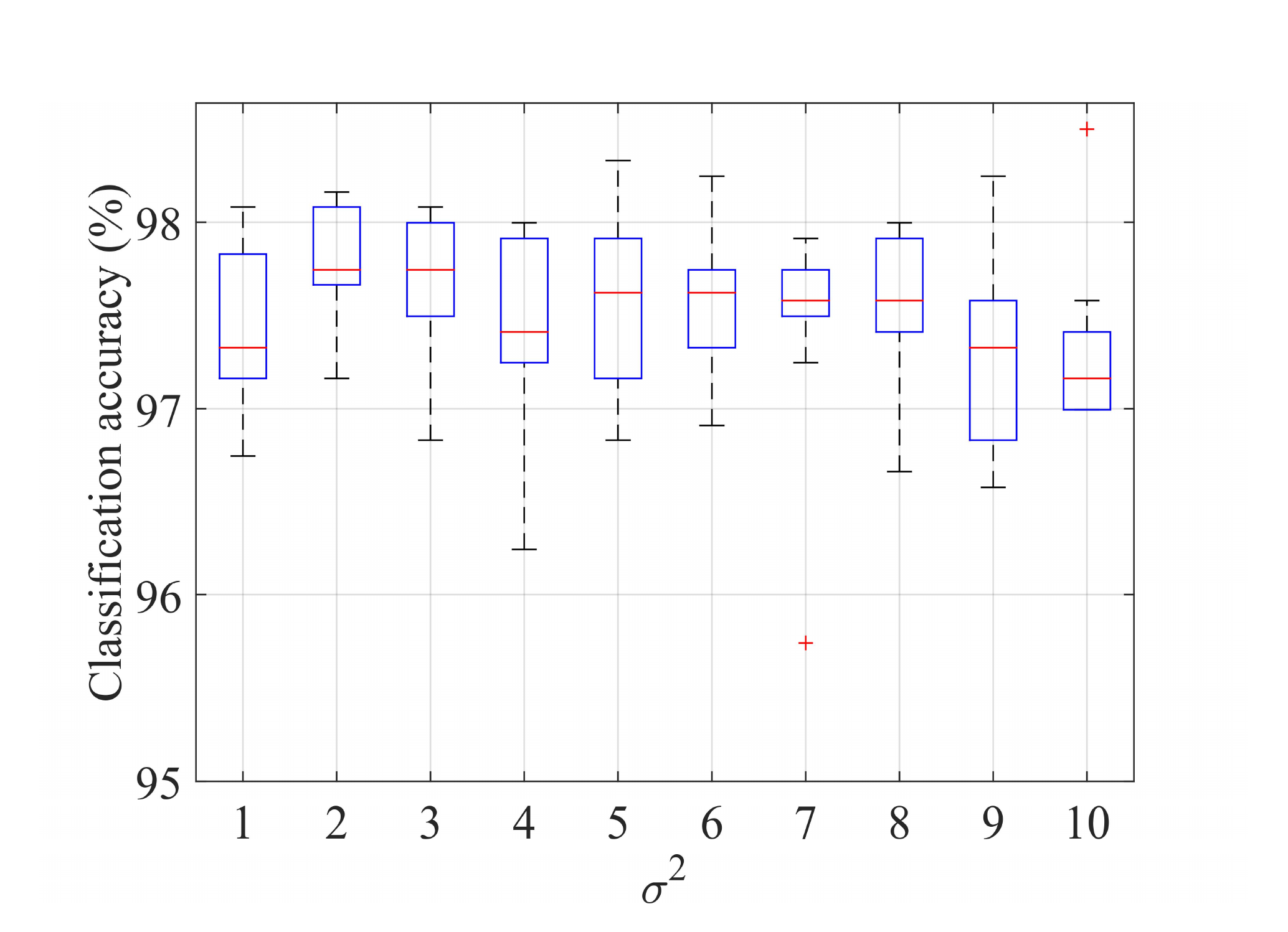}\label{Fig:Sigmasep}}
  \subfigure[]{\includegraphics[width=0.23\textwidth]{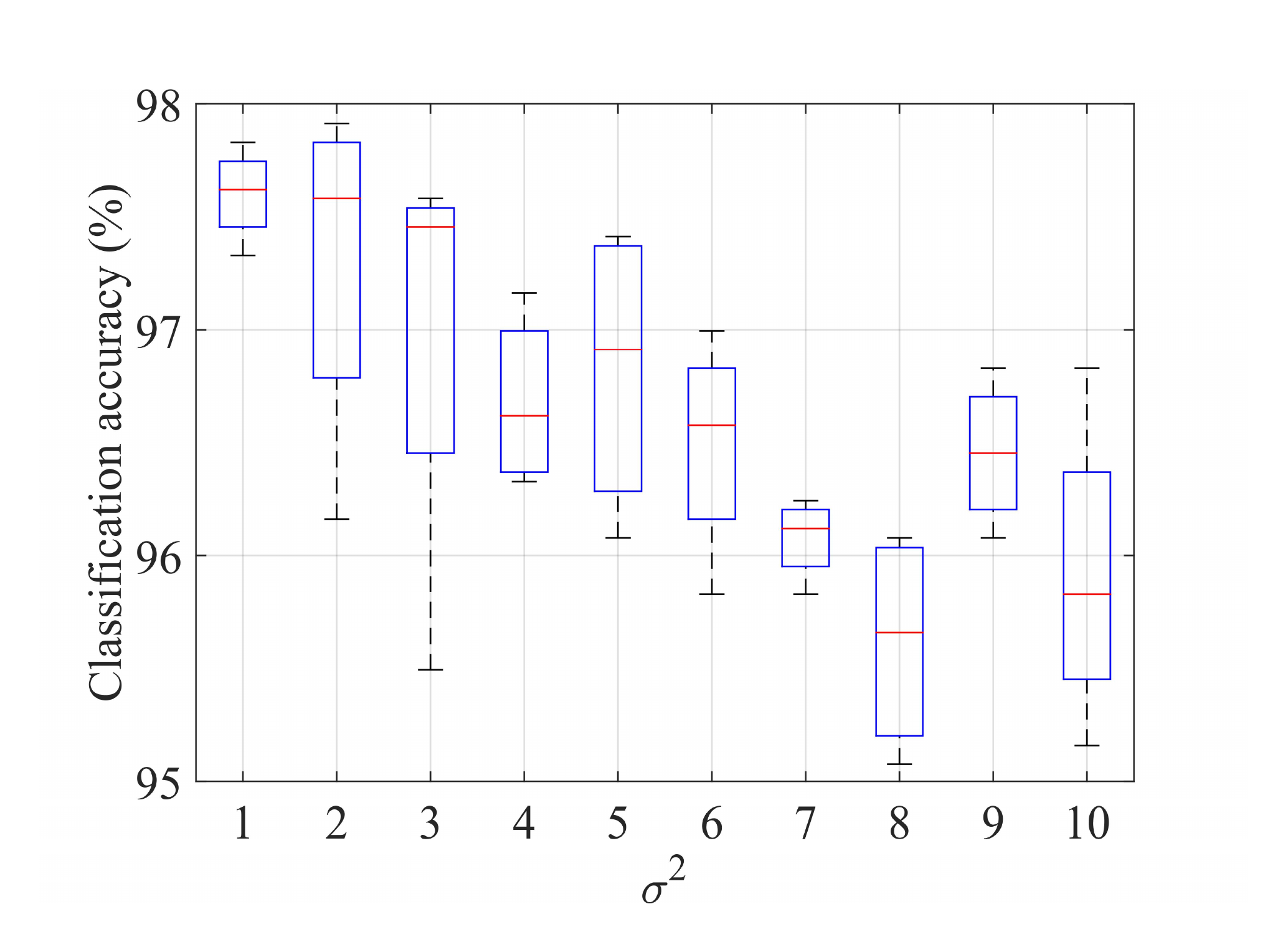}\label{Fig:SigmaNonsep}}
  \caption{Effects of $\sigma$ for two schemes on the classification accuracy. (a). Sep-DNAOL. (b). NonSep-DNAOL.}\label{Fig:Parasigma}
\end{figure}

 In the following experiments, we will evaluate the initialization of $\mathbf{A}\sim \mathcal{N}(\mathbf{0},\sigma^2 \mathbf{I})$ and the performances are illustrated in Figs. \ref{Fig:Parasigma}. According to the results, we can see that NonSep-DNAOL will be more sensitive to $\sigma$ and its classification accuracy varies from almost $98\%$ to $95.8\%$ while that of Sep-DNAOL will keep stable between $97\%$ and $98\%$.

\subsection{Framework Evaluation}

\par Finally, we will compare our two DNAOL frameworks with some typical DL frameworks based on linear synthesis model on different image databases to demonstrate their effectiveness and superiorities. To meet the goal of this paper, we will mainly concern the time cost in the training and testing phases and the corresponding classification accuracy to evaludate the performances of different algorithms. We may declare that since no more discrimination promoting strategies have been involved in our model, effectiveness and efficiency would only be fairly validated by comparing Sep-DNAOL and NonSep-DNAOL with DLSI \cite{ramirez2010classification}, DPL \cite{Gu2014} and with D-KSVD \cite{zhang2010discriminative}, respectively. Even so, we will still list some following state of the art frameworks with linear synthesis model to show the superiorities. Sparse representation classifier (SRC) and collaborative representation classifier (CRC) are two baseline synthesis models, where the training samples are directly served as the model parameters, namely dictionary. The main difference between these two approaches appears in the distinct regularizations on the features, where $\ell_1$ norm is exploited in SRC and square $\ell_2$ norm is adopted in CRC. Additionally, we also compare Sep-DNAOL with the novel dictionary learning approach of Fisher discriminative dictionary learning (FDDL) \cite{yang2014sparse}, where some more discriminative structure constraint or regularizations are involved to enhance the classification accuracy. Besides, NonSep-DNAOL will be further compared with the similar approaches of Label consistent KSVD (LC-KSVD2) \cite{jiang2013label} and joint embedding and dictionary learning (JEDL) \cite{Zhang2016}. The following comparison results reported are all quoted from the corresponding papers or simulated with the optimal settings \cite{jiang2013label,Gu2014,Zhang2016}.
\subsubsection{Datasets and protocols}
\begin{itemize}
  \item \textbf{Extended Yale B of human faces (E-YaleB)} \cite{Georghiades2001}: This database contains 2414 frontal face images from 38 classes where about 64
images taken under different illumination conditions and expressions will be collected for each person. Some sample images are shown in Fig. 6(a).  Following the common setting, the random features from a 504 dimensional input domain will be exploited to evaluate the algorithms \cite{jiang2013label} and each input feature vector will be firstly normalized with the unit $\ell_2$ norm. For this database, 32 samples per class will be randomly picked as the training samples while the rest are used as the testing samples. For our two frameworks, $\alpha=1\times 10^{-4}$, $\tau=7\times 10^{-6}$ and $\sigma^2=5$ are set in Sep-DNAOL while $\tau=9\times 10^{-5}$ $\sigma^2=1$ in NonSep-DNAOL and the equivalent feature dimension will be set as $1040~(30\times 38)$.
  \item \textbf{AR database of human faces (AR)} \cite{ding2010features}: This face database will include more occlusion images such as wearing sunglass and scarf and we select some normal sample images shown in Fig. 6(b). There are 2,600 images of 100 classes, and 20 images of each class are used for training and 6 images are remained for testing following the common setting. For this database, the 540 dimensional random feature vectors will be utilized as the input samples and the equivalent feature dimension will be set as $2000~(20\times 100)$. $\alpha=1\times 10^{-3}$, $\tau=1\times 10^{-4}$, $\sigma^2=4$ in Sep-DNAOL while $\tau=3\times 10^{-4}$, $\sigma^2=2$ in NonSep-DNAOL.
  \item \textbf{Caltech-101 objects dataset (Caltech-101)} \cite{fei2007learning}: Caltech-101 database contains 9144 images from 102 categories including 101 object classes and a background class and some sample images are shown in Figs. 7(a) and 7(b). Following the common setting, 30 samples from each category of Caltech-101 are used for training and the rests are for testing, where the feature dimension will be set as $3060~(30\times 102)$. For this database, spatial pyramid matching and bag-of-words framework are used for coarse feature extraction, where SIFT features are extracted for three grids of size $1\times 1$, $2\times 2$ and $4\times 4$. Then the standard vector quantization method is used to pool the SIFT features, where max pooling strategy is adopted. These resulted features are further reduced to 3000 dimension by means of principal component analysis as the final input samples. We set $\alpha=7\times 10^{-4}$, $\tau=9\times 10^{-4}$ and $\sigma^2=80$ in Sep-DNAOL while $\tau=0.01$ and $\sigma^2=20$ in NonSep-DNAOL.
  \item \textbf{15-Scene database (15-scene)} \cite{lazebnik2006beyond}: For the 15-scene database, it collects images from 15 scenes shown in Figs. 7(c) and 7(d). 100 samples of each class are employed for training, and the input samples are of 3000 dimensionality which are obtained according to the same preprocess of Caltech-101 similarly. However, the feature dimension this database is set to be $450~(30\times 15)$ following the common setting and thus it performs the classification in a lower dimensional feature domain. $\alpha=1\time 10^{-5}$, $\tau=3\times 10^{-3}$ and $\sigma^2=80$ in Sep-DNAOL, and $\tau=3\times 10^{-3}$, $\sigma^2=20$ in NonSep-DNAOL, respectively.
\end{itemize}
\begin{figure}
  \centering
  \includegraphics[width=0.5\textwidth]{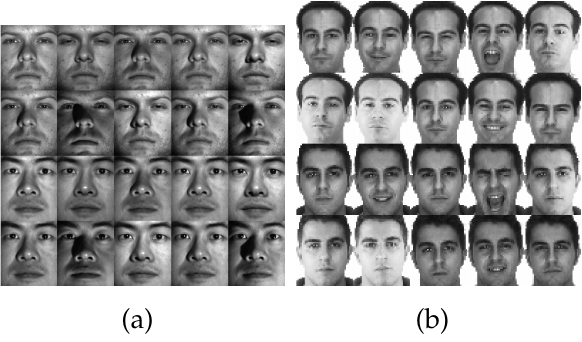}\\
  \caption{Sample images from EYaleB and AR. (a). EYaleB. (b). AR.}\label{Fig_SampleImage_Face}
\end{figure}
\begin{figure}
  \centering
  \includegraphics[width=0.5\textwidth]{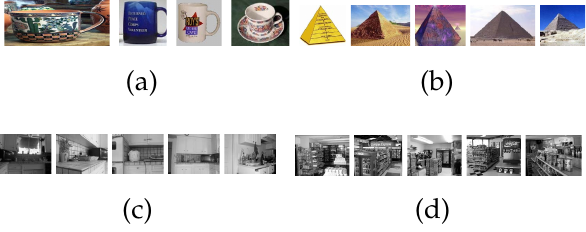}\\
  \caption{Sample images from Caltech-101 database and scene-15 database. (a)(b). Cup, Pyramid. (c)(d) Kitchen, Store.}\label{Fig_SampleImage_Caltech101andscene}
\end{figure}

\begin{table}\caption{Overall Accuracies ($\%$) Comparison and Best Results are Bolded}\label{Tab:OA}  \centering
  \begin{tabular}{|c|c|c|c|c|c|}
\hline
 Type& Algorithms & E-YaleB & AR & Caltech101 & 15-scene\tabularnewline
\hline
\hline
\multirow{7}{*}{1} & Sep-DNAOL &\textbf{97.9}  & \textbf{98.5} &71.8  &\textbf{98.2} \tabularnewline
\cline{2-6}
 & SRC &96.8  & 97.5 & 70.7 & 91.8\tabularnewline
\cline{2-6}
 & CRC & 97.0 &97.7  & 70.4 & 95.2\tabularnewline
\cline{2-6}
 & DLSI & 97.0 &97.5  &73.1  & 92.5\tabularnewline
\cline{2-6}
 & FDDL & 96.7 &97.5  & \textbf{73.2} & 97.6\tabularnewline
\cline{2-6}
 & DPL & 97.5 & 98.3 &71.2  & 97.7\tabularnewline
\hline
\hline
\multirow{3}{*}{2} & NonSep-DNAOL &\textbf{97.8}  & \textbf{98.6} &73.1  & \textbf{97.9}\tabularnewline
\cline{2-6}
 & JEDL & 95.4 & 98.3 &\textbf{74.4}  &95.7 \tabularnewline
\cline{2-6}
 & LC-KSVD2 &96.7  & 97.8 & 73.6 &92.9 \tabularnewline
 \cline{2-6}
 & D-KSVD & 95.4 & 95.0& 73.0 &89.1 \tabularnewline
 \hline
\end{tabular}

\end{table}

\subsubsection{Overall accuracies comparison}
\par Let us firstly focus our sight on the overall accuracies (OA) summarized in Table \ref{Tab:OA}, which is divided into two parts according to their different implementations. Firstly, considering the implementation 1, a separable dictionary comprising class-specific atoms are learned or directly utilized from the training samples in each comparison algorithm. We can see that the proposed Sep-DNAOL will outperform the compared ones on three databases except Caltech-101. Note also from the results that Sep-DNAOL does achieve an evidently better performance on all databases than its twin brother algorithm with the almost same strategies, namely DLSI. As a straightforward conclusion, when the cosparse model meets the classification task, the proposed NACM will indeed appear more discrimination than the conventional linear synthesis model. Compared with DPL, Sep-DNAOL can also bring about higher accuracies because the feature extractor in training and testing phases are still inconsistent in DPL. Compared with the other state-of-the-art DL algorithms, although many sophisticated discrimination promoting strategies have been involved in their frameworks, e.g., FDDL, they only achieve a marginally better performance on Caltech-101 database than DPL and our Sep-DNAOL. We conjecture that it is caused by the following two folds. On one hand, the proposed model is more sensitive to the initializations than the generative ones because we can produce any features via $\mathbf{f}=\mathcal{S}_{\mathbf{\Lambda}}(\mathbf{Ax})$ while linear synthesis model will restrain $\mathbf{x}$ to be synthesized by the dictionary $\mathbf{D}$ and $\mathbf{f}$ as $\mathbf{x=Df}$. In our model, the random matrices drawn from the Gaussian distribution are used as the initialization while the most compared algorithms use the off the shelf training sample matrices or pre-trained dictionary with K-SVD \cite{aharon2006svd}. On the other hand, images in this database are appearing large variations, the amount of the exploited training samples is intrinsically deficient for a discriminative model to characterize $p(\mathbf{y},\mathbf{f}|\mathbf{x})$ in this case. As a conclusion, the discriminative model will generally achieve a higher classification performance than the generative model, but the limited labeled samples will conversely degrade its performance than generative ones \cite{Bernardo2007}. Nevertheless, we will later see from the time costs comparison that the algorithms with the generative model will pay the price of heavily increased computational burdens during the training phase. Likewise, the same conclusions can be derived from the performances of the type 2 implementation, where a non-separable dictionary is learned in the compared frameworks. In this scenario, the proposed NonSep-DNAOL significantly improves the classification accuracies than those compared generative models on three databases, especially its counterpart algorithm of D-KSVD \cite{zhang2010discriminative}. In a summary, owning to the proposed NACM, two frameworks of Sep-DNAOL and NonSep-DNAOL can always generate a better classification accuracies than their generative counterparts with the same discriminative strategies. As a consequence, it will appear more potentials when it is devoted to the classification tasks with the sufficient training samples. In the next part, another remarkable superiority of low computational cost will be validated.
\subsubsection{Time costs comparison} Despite the classification accuracies, we will finally evaluate the time costs in training and testing phases of classifying one query sample for different algorithms.
\par
\begin{figure}
  \centering
  \subfigure[]{\includegraphics[width=0.23\textwidth]{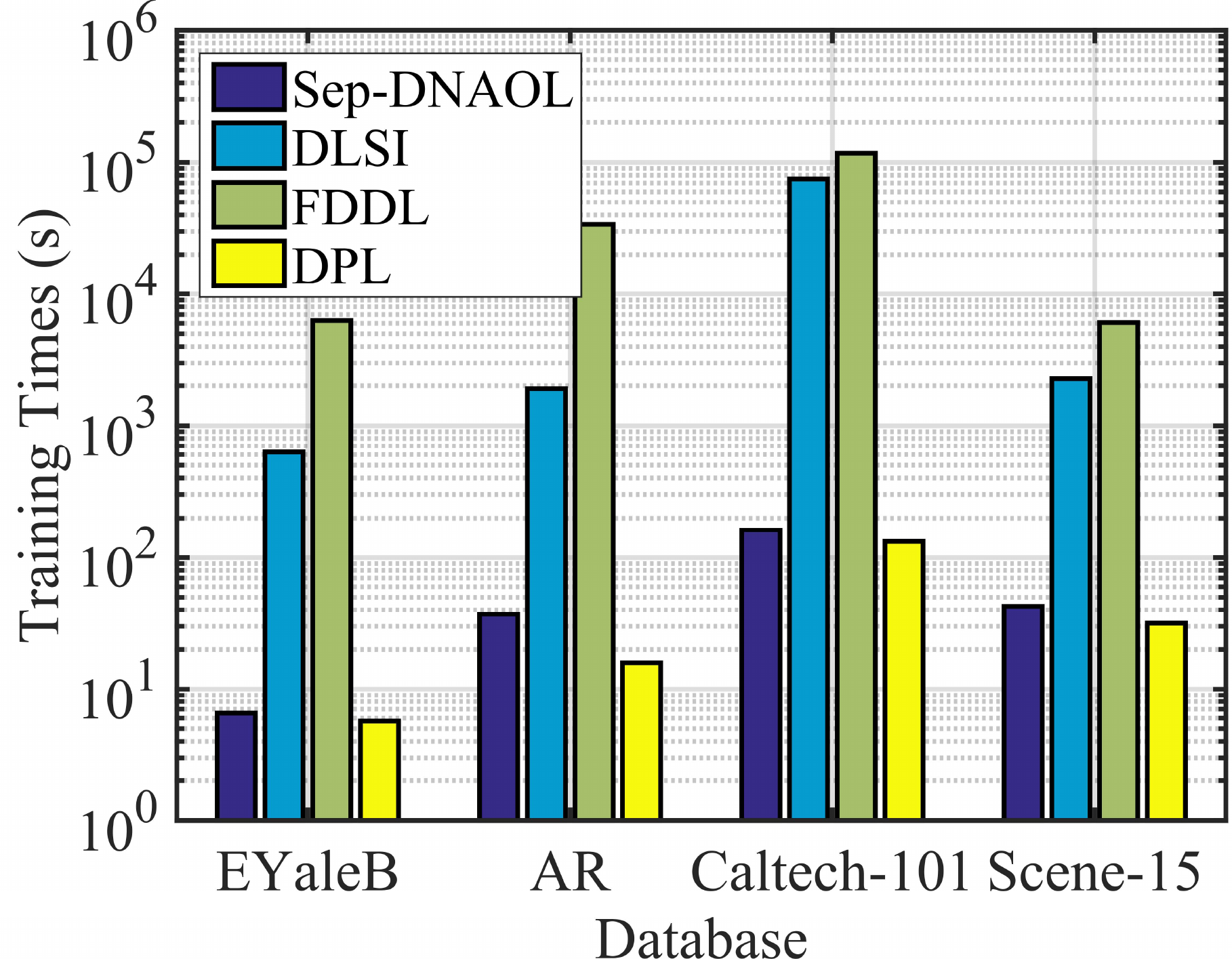}\label{Fig:Septime}}
  \subfigure[]{\includegraphics[width=0.23\textwidth]{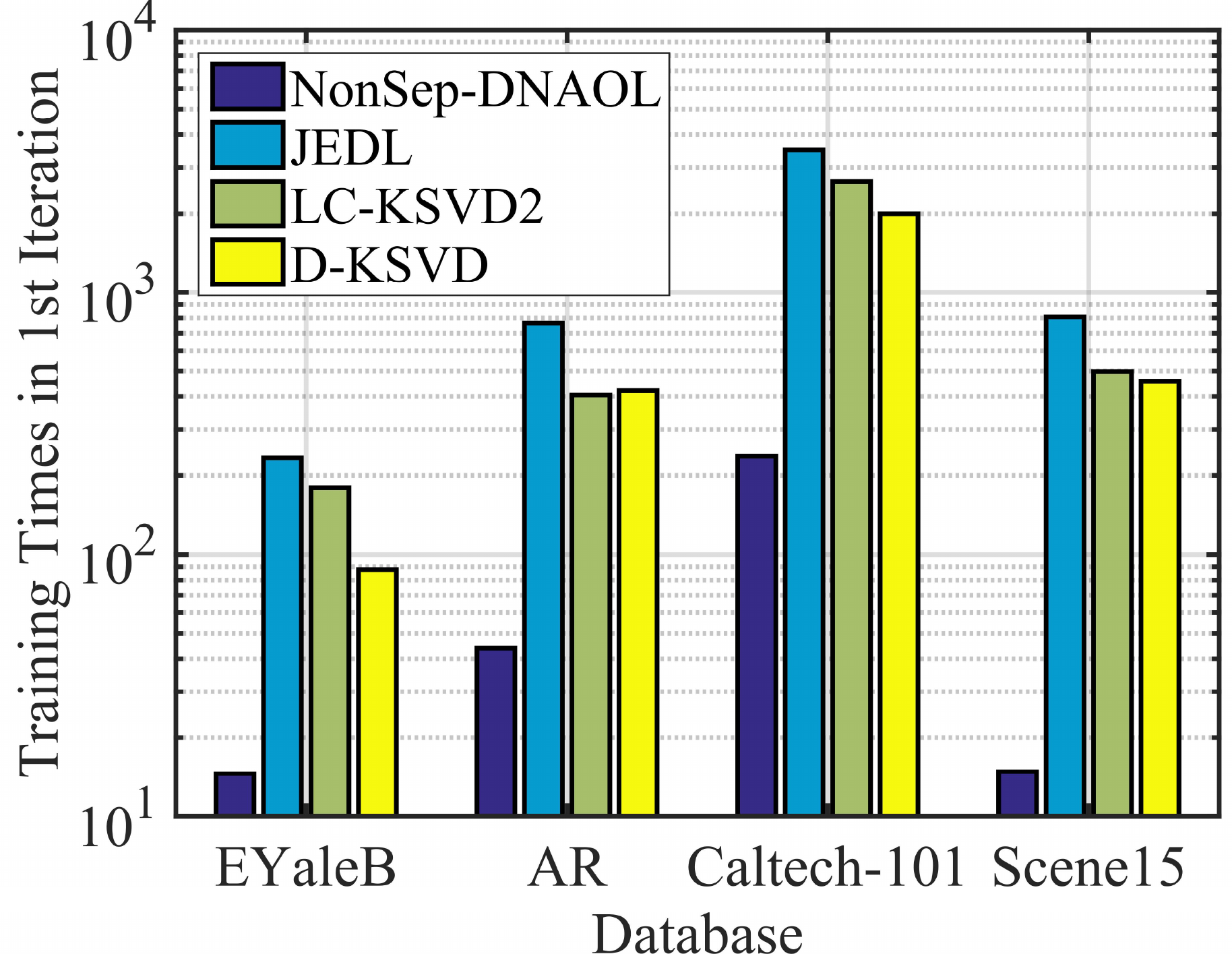}\label{Fig:NonSeptime}}
  \caption{Time costs comparison in training phase. (a).first type of implementation. (b). second type of implementation.}\label{Fig:Trainingtime}
\end{figure}

 Let us first examine the training time costs of those algorithms for the first type implementation, where several class-specific model parameters are learned. Since SRC and CRC directly exploit the input samples as the model parameters without learning, these two algorithms will not be compared. For the rest of four frameworks, Sep-DNAOL, DLSI, FDDL and DPL, we will compute their average overall training times and plot the results in Fig. \ref{Fig:Septime} in the order of magnitudes approximately. At the first sight of the plot, we can obviously conclude that two discriminative models, Sep-DNAOL and DPL will spend much less time on training model, where they are roughly 10 or 100 times faster than two generative models. We can also conclude from comparing the time costs on AR, Caltech-101 and Scene-15 that an increased number of categories will aggravate more cost in training phase than dimension increment of the input sample for this type of implementation but the training time for two discriminative models are still acceptable allowing their potential applications to the large scaled problem, where less than 200s will be required for  Caltech-101. On the contrary, for two generative models, it will spend about $1\times 10^4$s on the training process let alone the time cost for parameters validation. This time-consuming process will heavily limit the applications of the generative dictionary learning algorithms for some large scaled classification tasks. Comparing Sep-DNAOL with DPL, our framework is slightly slower than DPL because a nonlinear feature selection operator is extra involved to generative a MAP estimation of features. Nevertheless, feature mapping is consistent in our model which can improve the classification accuracy than DPL. In addition to the first type, we turn to those algorithms with the second type of implementation including NonSep-DNAOL, LC-KSVD2, JEDL and D-KSVD. Since these three compared algorithms all exploit K-SVD for dictionary learning, the average time cost in the first iteration including initialization process will be evaluated for simplicity and the results are plotted in Fig. \ref{Fig:NonSeptime}. The same conclusion can be obtained from the results that NonSep-DNAOL reduces a much fewer time cost than other three compared algorithms by means of typical K-SVD for dictionary update.
\par
\begin{figure}
  \centering
  \subfigure[]{\includegraphics[width=0.23\textwidth]{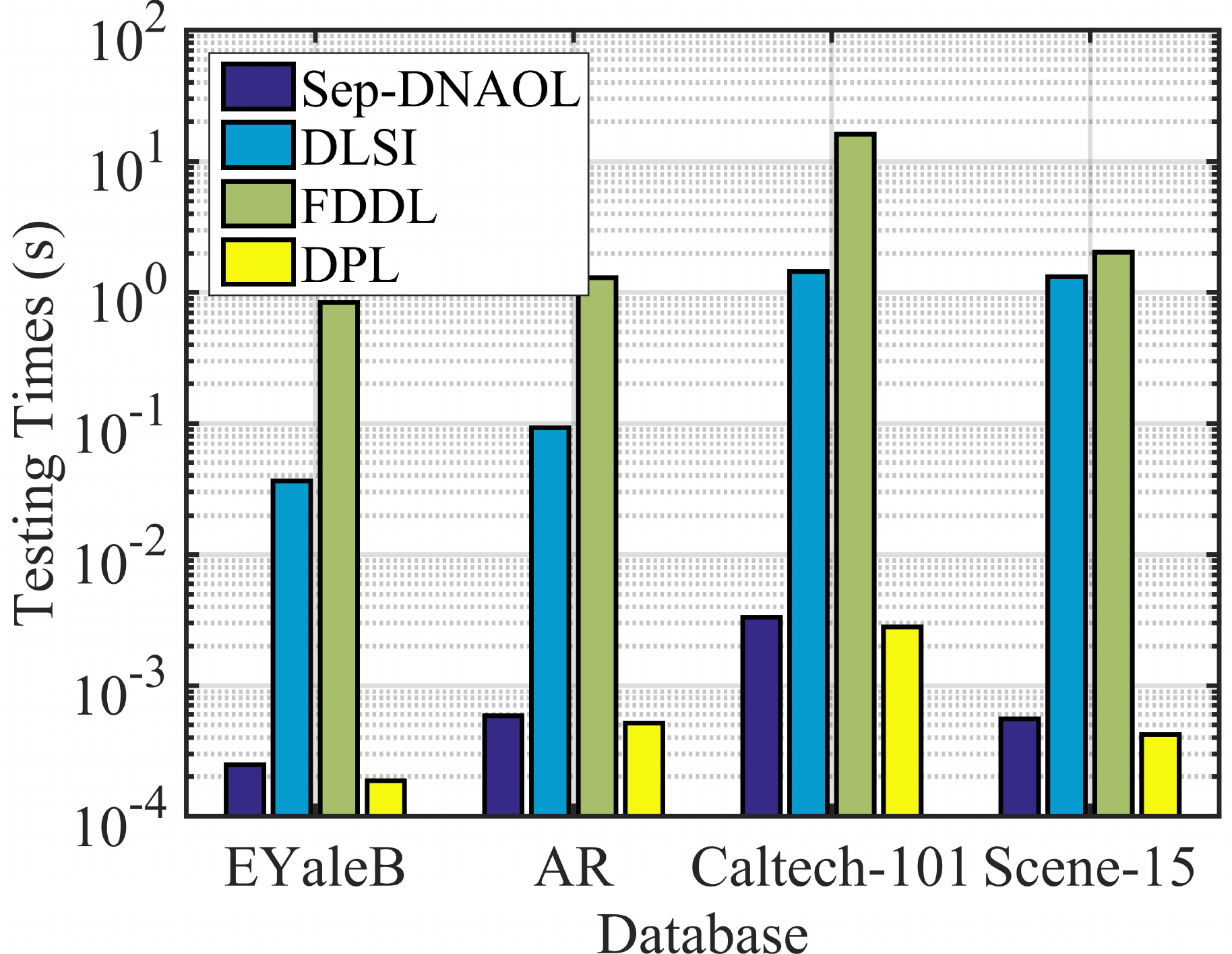}}
  \subfigure[]{\includegraphics[width=0.23\textwidth]{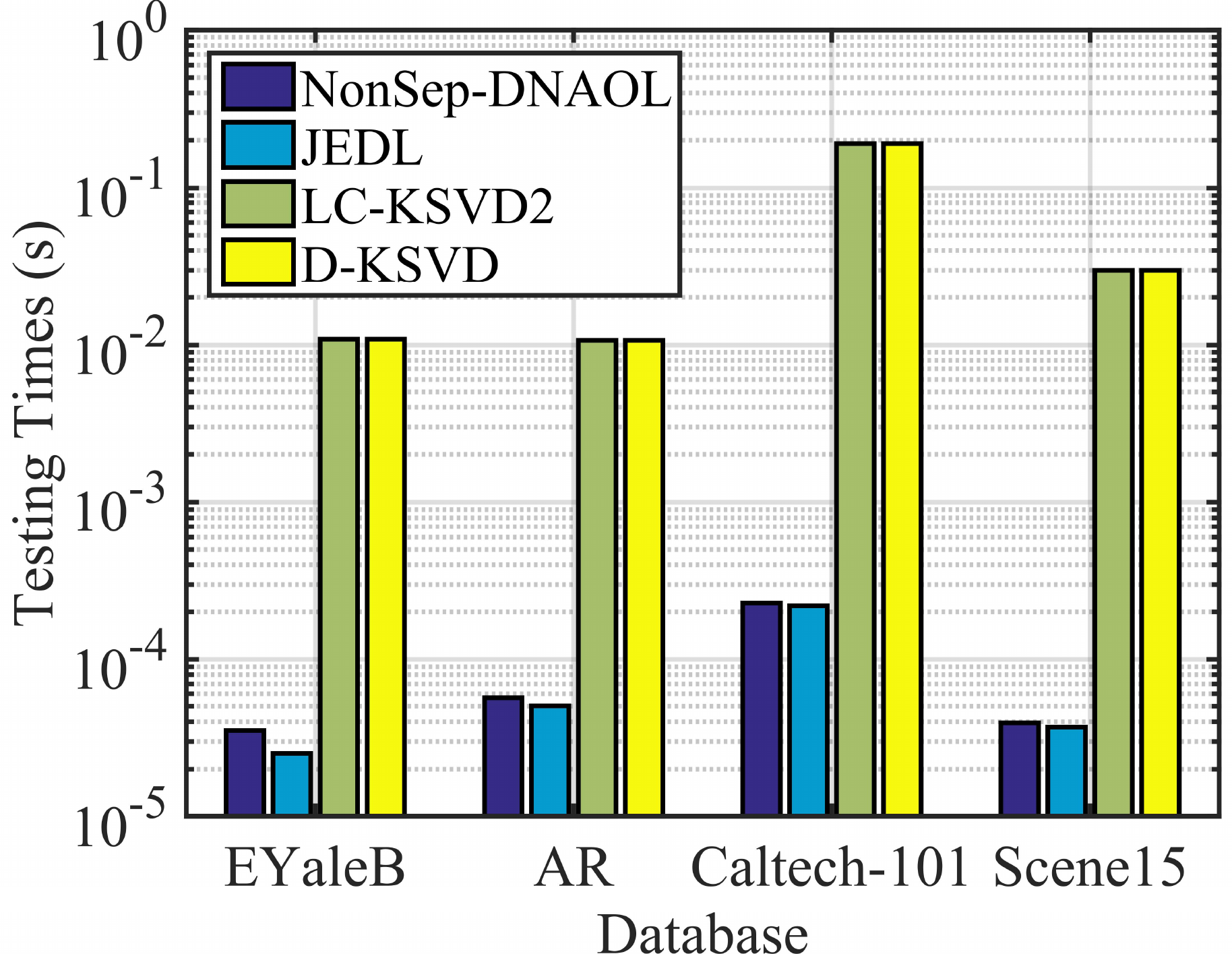}}
  \caption{Time costs comparison in testing phase. (a).first type of implementation. (b). second type of implementation.}\label{Fig:testtime}
\end{figure}

 Finally, we will check the testing times in classifying one query sample for different algorithms shown in Figs. \ref{Fig:testtime}. Totally speaking, the main difference between our two schemes and other compared algorithm focuses on the cost on extracting the feature vector. More specially, our two schemes together with JEDL and DPL will directly generate the feature vector of a query sample with the closed form linear or nonlinear mapping as $\mathbf{f}=\mathcal{P}(\mathbf{x})$ while the others should solve a complicated inverse problem. It follows that the results in Figs. \ref{Fig:testtime} show that the former mentioned four algorithms will cost fewer testing time in classifying one sample than the conventional generative frameworks.
\section{Concluding Remarks}\label{Sec:Concluding}
This paper presents a nonlinear analysis cosparse model (NACM) to allow learning a task oriented discriminative feature transformation and regularization simultaneously. When this cosparse model meets image classification, it is served as a parametric feature model in a novel discriminative nonlinear analysis operator learning framework (DNAOL), which successfully solves the deficiencies of the conventional regularized linear synthesis model by characterizing the posterior distribution. Evaluated on four image benchmark databases, DNAOL will not only achieve the better or at least competitive classification accuracies than the state-of-the-art algorithms but it can also significantly reduce the time complexities in both training and testing phases. As a general model, NACM can be potentially devoted to other tasks and it can be readily served as a basic building block to develop a hierarchical feature model. The main deficiencies of DNAOL can be empirically observed that it is more sensitive to the initializations than the generative model and the classification performance will be degraded as the number of training samples decreases. Future work will attempt to address these issues by developing a hybrid discriminative and generative model.

\bibliographystyle{IEEEtran}
\bibliography{IEEEabrv,egbib}

\end{document}